\documentclass[10pt,twocolumn,letterpaper]{article}

\usepackage[ruled, vlined]{algorithm2e}

\usepackage{cvpr}
\usepackage{epsfig}
\usepackage{times}
\usepackage{graphicx}

\usepackage{amsmath}
\usepackage{amssymb}
\usepackage{amsthm}
\usepackage{booktabs}
\usepackage{microtype}
\usepackage[pagebackref=true,breaklinks=true,letterpaper=true,colorlinks]{hyperref}

\newcommand{\citet}[2]{#1\,\cite{#2}}
\newcommand{\citep}{\cite}

\newtheorem{prop}{Proposition}
\newtheorem{probrem}{Problem}

\newcommand{\Sec}[1]{{Sec.\,\ref{#1}}}
\newcommand{\Circ}{\mathrm{Circ}}

\cvprfinalcopy
\def\cvprPaperID{6679}

\ifcvprfinal\fi

\begin{document}

\title{On the Structural Sensitivity of Deep Convolutional Networks\\to the Directions of Fourier Basis Functions}
\author{
    Yusuke Tsuzuku$^{1,2}$, \,\,
    Issei Sato$^{1,2}$ \vspace{3pt}\\
    $^1$The University of Tokyo, $^2$RIKEN\\
    {\tt\small tsuzuku@ms.k.u-tokyo.ac.jp}, \hspace{3pt} {\tt\small sato@k.u-tokyo.ac.jp}
}
\maketitle

\begin{abstract}
  Data-agnostic quasi-imperceptible perturbations on inputs are known to degrade recognition accuracy of deep convolutional networks severely.
  This phenomenon is considered to be a potential security issue.
  Moreover, some results on statistical generalization guarantees indicate that the phenomenon can be a key to improve the networks' generalization.
  However, the characteristics of the shared directions of such harmful perturbations remain unknown.
  Our primal finding is that convolutional networks are sensitive to the directions of Fourier basis functions.
  We derived the property by specializing a hypothesis of the cause of the sensitivity, known as the linearity of neural networks, to convolutional networks and empirically validated it.
  As a by-product of the analysis, we propose an algorithm to create shift-invariant universal adversarial perturbations available in black-box settings.
\end{abstract}

\section{Introduction}
  \noindent
  Malicious perturbations on inputs can easily change predictions of deep learning models\,\citep{Intriguing}.
  These perturbations are called adversarial perturbations or adversarial examples.
  They have been intensively studied concerning deep convolutional networks for object recognition tasks \citep{CW,FGSM,AtScale,DeepFool,Intriguing,LMT}.
  They are attracting attention because they are potential security issues.
  One of the intriguing aspects of adversarial perturbations is their universality.
  \citet{Szegedy \etal}{Intriguing} observed transferability of the perturbations between classifiers.
  \citet{Papernot \etal}{Transferability,PracticalBlackBox}
  exploited the transferability to attack black-box models.
  Some adversarial perturbations transfer not only between classifiers but also between inputs.
  \citet{Goodfellow \etal}{FGSM} first discovered the universality, and
  \citet{Moosavi{-}Dezfooli \etal}{Universal} studied this phenomenon in more detail.
  They found that a single perturbation can change models' predictions for a significant portion of data points.
  Such input-agnostic perturbations are called universal adversarial perturbations (UAPs).
  The perturbations also generalize between different networks to some extent.

  \begin{figure}[t]
    \center
    \begin{minipage}{.9\hsize}
      \center
      \begin{minipage}{.49\hsize}
        \center
        \includegraphics[width=\columnwidth]{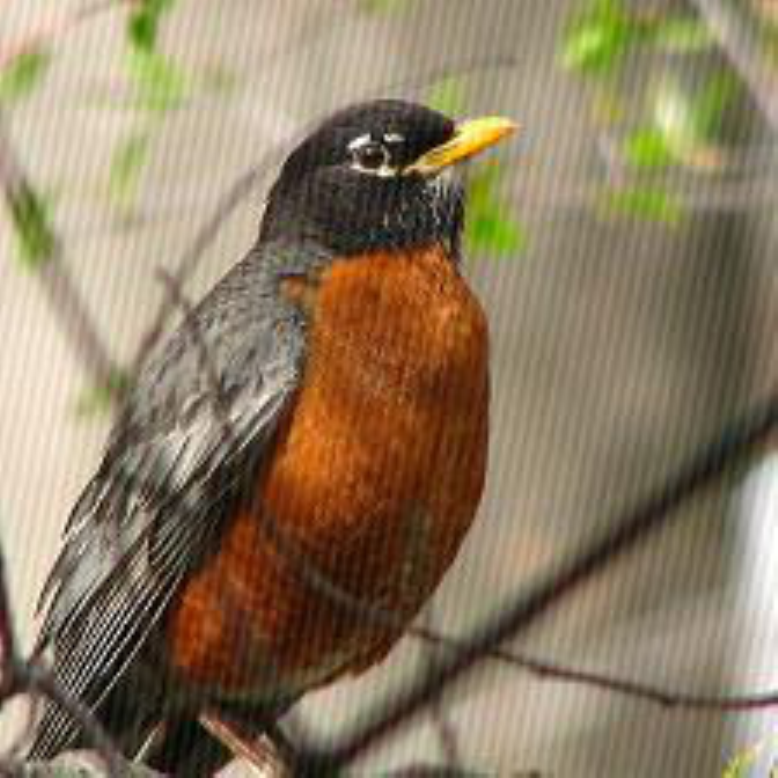}
      \end{minipage}
      \begin{minipage}{.49\hsize}
        \center
        \includegraphics[width=\columnwidth]{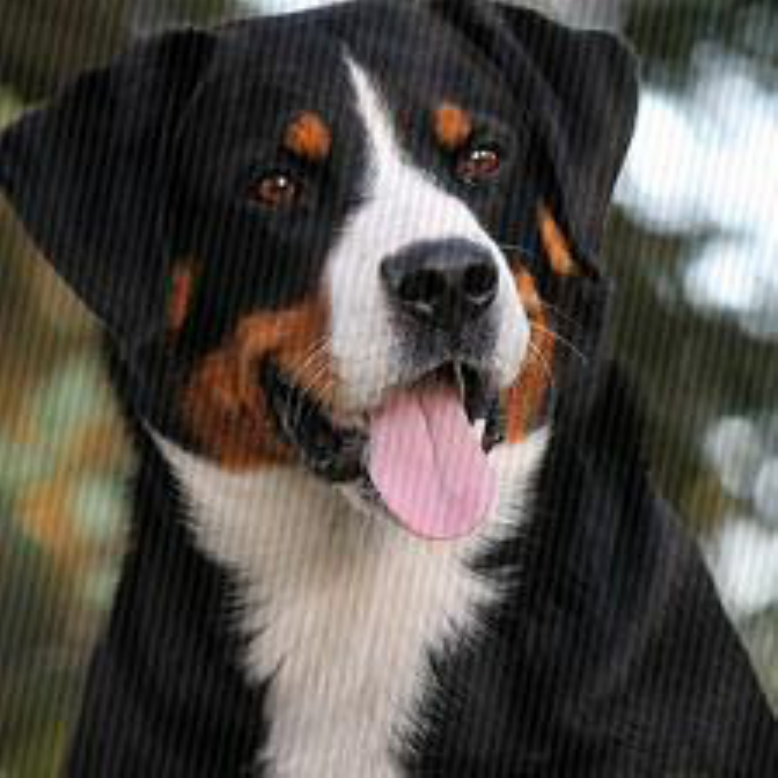}
      \end{minipage}
    \end{minipage}
    \begin{minipage}{.9\hsize}
      \center
      \begin{minipage}{.49\hsize}
        \center
        \includegraphics[width=\columnwidth]{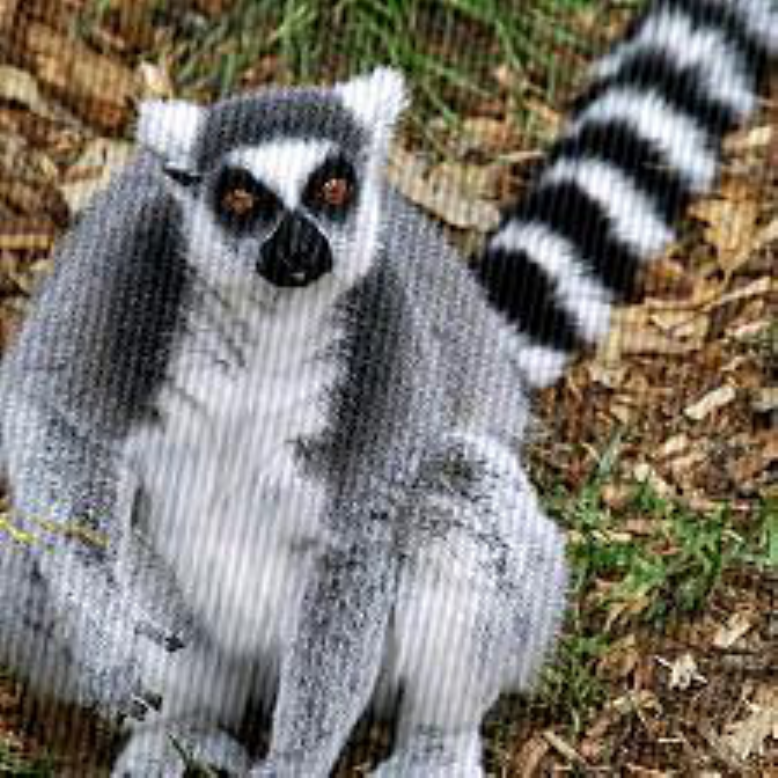}
      \end{minipage}
      \begin{minipage}{.49\hsize}
        \center
        \includegraphics[width=\columnwidth]{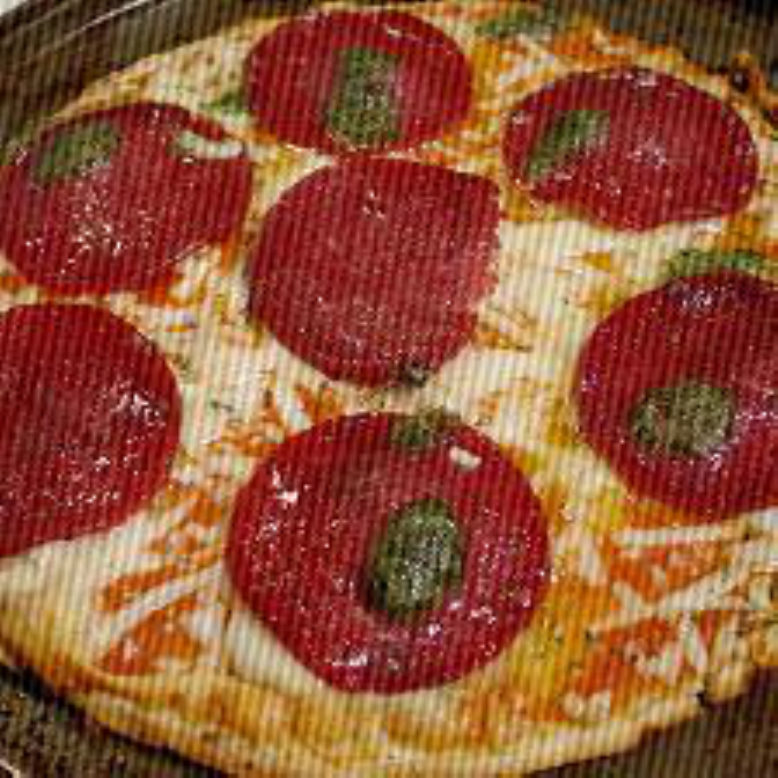}
      \end{minipage}
    \end{minipage}
    \caption{
      Examples of images perturbed by single Fourier attack.
      Added perturbation is the same as in Figure\,\ref{fig:hffm_noise_examples}.
      The size of perturbations is $10/255$ in $\ell_\infty$-distance for the first row and
      $20/255$ for the second.
      In Sec.\,\ref{subsec:fool_ratio}, we show that the single $10/255$ and $20/255$ perturbations could change predictions for around $40\%$ and $70\%$ of inputs for various architectures, respectively.
    }
    \label{fig:hffm_examples}
  \end{figure}

  \begin{figure*}[t]
    \center
    \includegraphics[width=\textwidth]{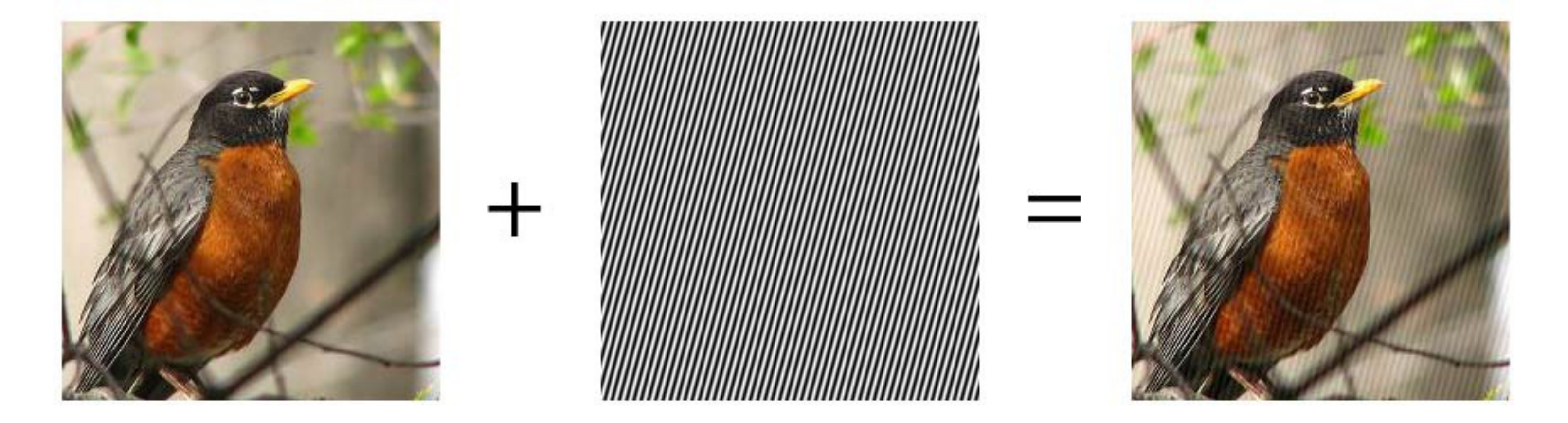}
    \caption{Illustration of our UAPs creation algorithm. We only tune frequency of noise. We do not need access to model parameters, output logits, or training data.}
  \end{figure*}

  We are primarily concerned with UAPs because of their relation to statistical generalization guarantees of deep learning models.
  For example, studies using PAC-Bayes\,\citep{PACBayesianSpectralMargin}, compression\,\citep{CompressionBound}, and minimum description length\,\citep{KeepNNSimple} are all concerned with how perturbation propagates networks.\footnote{
    In these studies, we consider perturbations on weights, not inputs.
    However, perturbations on weights become noises on inputs of the network's subnetworks and investigating UAPs is still useful.
  }
  In these analyses, how each perturbation changes accuracy on training data and true data distribution matters.
  In other words, we have an interest in perturbations transferable between inputs.
  These are nothing else but UAPs.
  We try to shed lights on the tendency of UAPs and how they propagate in convolutional neural networks.

  Several prior studies tried to understand the properties of the universality and transferability of adversarial perturbations.
  \citet{Goodfellow \etal}{FGSM} explained the existence of adversarial examples, their transferability, and their universality using linearity of deep neural networks.
  \citet{Tram\`er \etal}{TransferableSpace} investigated the transferable subspace of adversarial perturbations and suggested that it will consist of a high-dimensional continuous subspace.
  \citet{Moosavi-Dezfooli \etal}{AnalysisUAP} showed that the existence of universal adversarial perturbations is inevitable given strong geometrical assumptions on the decision boundaries of models.

  Given the transferability and the universality of adversarial perturbations, it is natural to expect the existence of a set of directions to which most networks are input-agnostically sensitive.
  If we can characterize such directions, it enables us to improve robustness against such perturbations in principled manners.
  Additionally, we may design better posteriors, weights, or compression algorithms to achieve empirically better generalization bounds.
  However, prior work can only generate such perturbations by sequential optimization and lacks their useful characterization.
  We provide a missing characterization of directions by analyzing Fourier basis functions.
 
  The motivation of our analysis comes from two parts.
  The first is the linear hypothesis of vulnerability, and the second is a property of linear convolutional layers that the singular vectors of which are Fourier basis functions.
  The property indicates that sensitive directions of convolutional networks are a combination of a few Fourier basis functions.
  Through extensive experiments on various architectures and datasets, we found networks are sensitive to the directions of Fourier basis functions of some specific frequencies.
  In other words, we could characterize at least a subset of universal and transferable adversarial perturbations through Fourier basis functions.
  We also observed that some adversarial perturbations exploit the sensitivity to Fourier basis functions.
  These findings not only provide a new characterization of adversarial perturbations with benefits described in the preceding paragraph but also suggest a possibility that some known properties of the universality of adversarial perturbations might be due to the structure of convolutional networks.

  As a by-product of our analysis, we also developed a method to create shift-invariant universal adversarial perturbations, which is available in black-box settings.
  Figure\,\ref{fig:hffm_examples} shows examples of perturbed images created by our algorithm, which is explained in Sec.\,\ref{sec:fourier_analysis}. Our perturbations have simple and shift-invariant patterns, yet achieved high fool ratio on various pairs of architectures and datasets.

  Our contributions are summarized below.
  \begin{enumerate}
    \item We characterized spaces UAPs lie using Fourier basis functions.
    \item We evaluated our hypothesis in extensive experiments.
    \item We proposed a black-box algorithm to create shift-invariant universal adversarial perturbations.
  \end{enumerate}

\section{Related work}

  \subsection{Adversarial perturbations}

    One of the most famous algorithms for creating adversarial perturbations is the fast gradient sign method (FGSM)\,\citep{FGSM}.
    Let $J(\theta, x, t)$ be a loss with parameter $\theta$, an input $x$, and a target label $t$.
    Then, FGSM uses $\epsilon\cdot\mathrm{Sign}\left(\nabla_xJ(\theta, x, t)\right)$ as the perturbation,
    where $\epsilon$ is a scaling parameter.
    Another popular approach is performing gradient ascent on some loss $J(\theta, x, t)$.
    Depending on the choice of the loss and the optimization methods, there are numerous variants for attacks\,\citep{CW,DeepFool}.
    Adversarial training\,\citep{FGSM} is a current effective countermeasure against adversarial perturbations.
    \citet{Kurakin \etal}{AtScale} conducted a large-scale study on adversarial training, and
    \citet{Tram{\`{e}}r \etal}{Ensemble} extensively studied the transferability for defended and undefended models.
    Evaluations of defense methods are notoriously difficult\,\citep{Obfuscated,DeepMindObfuscated}.
    Thus, some studies have provided theoretically grounded defense methods\,\citep{OuterPolytope,FastReLU}.

  \subsection{Universal adversarial perturbations}
    \citet{Moosavi{-}Dezfooli \etal}{Universal} showed that some input-independent perturbations
    can significantly degrade classifiers' prediction accuracy.
    Such perturbations are called universal adversarial perturbations (UAPs).
    \citet{Moosavi{-}Dezfooli \etal}{Universal} created UAPs by sequentially optimizing perturbations until we achieve the desired fool ratio.
    During the creation, they did not need access to test data.
    They showed that UAPs could change over $80\%$ of the predictions of various networks trained on ILSVRC2012\,\citep{ILSVRC}.
    UAPs also generalize between network architectures to some extent.
    Recently, \citet{Mopuri \etal}{FastFeatureFool} and \citet{Khrulkov \etal}{ArtSingular} proposed activation-maximization approaches for the creation of UAPs.
    UAPs degrade the average performance of systems and have different nature from other kinds of adversarial examples.

  \subsection{Analysis of transferability and universality}
    \citet{Goodfellow \etal}{FGSM} explained the existence of adversarial examples, their transferability, and their universality by linear hypothesis.
    In their explanation, the directions of perturbations are the most important in adversarial examples.
    The hypothesis is based on the following three factors:
    (1) modern networks behave like linear classifiers,
    (2) adversarial perturbations are aligned with the weight vectors of models,
    (3) different models learn similar functions.
    Thus, adversarial perturbations generalize between clean examples, and also different models.
    \citet{Tram\`er \etal}{TransferableSpace} analyzed the dimensionality of the subspace that adversarial examples lie in.
    Using first-order approximation, they found that adversarial examples lie in a high-dimensional subspace, suggesting overwrap of the subspace between classifiers.
    However, the structure of the subspace is unknown except for its estimated dimensionality.
    \citet{Moosavi-Dezfooli \etal}{AnalysisUAP} analyzed the existence of UAPs using strong geometrical assumptions.
    They also proposed an algorithm to find UAPs using Hessian on input, while it is prohibitively slow with large inputs.

    We explain the existence of UAPs on the basis of the linear hypothesis of \citet{Goodfellow \etal}{FGSM}.
    We push forward the analysis concerning convolutional networks.

  \subsection{Fourier basis}
    \citet{Jo and Bengio}{MeasuringTendency} examined whether CNNs learn high-level features by using Fourier features.
    Some prior work used eps compression or other transformations as defenses against adversarial examples\,\citep{JPG,InputTransform,Saak}.
    They remove high-frequency features from images and relates to this paper.
    However, connections to universality have not been explored.
    Also, the effects of each frequency have not been studied.
    In a later section (\ref{sec:experiment}), our experiments show that adversarial perturbations do not necessarily lie in high-frequency spots.

\section{Preliminary}

  In this section, we describe the relationship between convolutional layers and Fourier basis.
  Notations are summarized in the supplementary material.

  \subsection{Fourier basis and discrete Fourier transformation}

    Let us define $\omega_N^{i,j} = \omega_N^{i}\omega_N^{j} \in \mathbb{C}^{N\times N}$, where $\omega_N=\exp(2\pi\sqrt{-1}/N)$ is the $N$-th root of an imaginary number.
    We define $F_N$ be a matrix such that colums are $n$ fourier basis functions with different frequencies.
    In other words, $F_N$ is a matrix such that
    \begin{equation}
      (F_N)_{u,v} = \frac{1}{\sqrt{N}}\omega_N^{u,v}.
    \end{equation}
    We notate the $i$-th row of $F_N$ as $(F_N)_i$.
    Let us define a transformation $S:\mathbb{C}^{N \times N}\rightarrow\mathbb{C}^{N \times N}$ as follows.
    \begin{equation}
      S(x)_{u,v} = \underset{m=0}{\overset{N-1}{\sum}}
      \underset{n=0}{\overset{N-1}{\sum}}
      x_{m,n}\exp(-2\pi \sqrt{-1}(um + vn)/N)
    \end{equation}
    This transformation $S$ is called discrete Fourier transformation (DFT).
    Both the transformation and its inverse can be calculated in the running time of $O(N\log N)$ by using fast Fourier transformation \citep{FFT}.

  \subsection{Decomposition of convolution operator}
    \label{subsec:decomposition}

    We define $Q_N := \frac{1}{N}F_N\otimes F_N$, where $\otimes$ is a Kronecker product.
    The eigenvectors of a doubly block circulant matrix are known to be $Q_N$ \citep{FundamentalsDIP}.
    Since $Q_N$ is unitary, a doubly block circulant matrix can be decomposed as $Q_NDQ_N^{\mathrm{H}}$, where $Q_N^{\mathrm{H}}$ is an adjoint matrix of $Q_N$, and $D$ is a complex diagonal matrix.
    In a case where channel size is one, since convolution is a doubly circulant matrix when the padding is ``wraps around''\citep{DeepLearning,SingularConv}, the above analysis is directly applicable.
    We can extend the result to multi-channel cases, i.e., $m \geq 1$.
    \begin{prop}
      \label{prop:decomposition}
      Let $M$ be a matrix which represents a convolutional layer with input channel size $m_{\mathrm{in}}$, output channel size $m_{\mathrm{out}}$, and input size $m_{\mathrm{in}}\times N \times N$.
      Then, $M$ can be decomposed as
      \begin{equation}
        M = \left( I_{m_{\mathrm{out}}} \otimes Q_N \right) L \left( I_{m_{\mathrm{in}}} \otimes Q_N \right)^{\mathrm{H}},
      \end{equation}
      where $L$ is a block matrix whose blocks are $N^2 \times N^2$ diagonal matrices.
    \end{prop}

\section{Fourier analysis}
  \label{sec:fourier_analysis}

  In this section, we show that the most sensitive direction of linear convolutional networks is a combination of a few Fourier basis functions.
  The analysis pushes forward the linear hypothesis of the cause of adversarial examples in \citet{Goodfellow \etal}{FGSM}.
  The linear approximation may not hold well for deep non-linear networks.
  However, we can still expect that adding some Fourier basis functions to inputs can largely disturb hidden representations of networks.
  We assume that the padding of convolutional layers are ``wraps around.''
  Notations are summarized in the supplementary material.
  Proofs of propositions are deferred to the supplementary material.

  \subsection{Sensitivity of stacked convolutional layers}

    We first consider stacked stride-$1$ convolutional layers without activation functions.
    In the case, we can show that the singular vectors of the whole layers can be represented by a linear combination of single Fourier basis functions between input channels.

    \begin{prop}
      \label{prop:stacked_conv}
      Let $M^{(i)}$ be a convolitional layer with input channel size $m^{(l)}$, output channel size $m^{(l+1)}$, input size $m^{(l)}\times N\times N$, and stride $1$.
      Let $M$ be a stacked convolutional layers with linear activation, i.e., $M(X) = \left( M^{(1)}\circ M^{(2)} \circ \dots \circ M^{(d)} \right) (X)$.
      Then, we can choose the right singular vectors of $M$ so that all of them can be represented by $a \otimes (F_{N})_i\otimes(F_{N})_j$ for some $i, j \in\{0,\dots,N-1\}$ and $a \in \mathbb{C}^{m^{(1)}}$.
    \end{prop}

    In other words, the most sensitive directions of linear convolutional neural networks without reduction layers is a single Fourier basis function.
    We can further extend the result to cases when there are normalization layers or skip connections.

    \begin{prop}
      \label{prop:skip_conv}
      Let $M^{(i)}$ be a convolitional layer with input channel size $m^{(l)}$, output channel size $m^{(l+1)}$, input size $m^{(l)}\times N\times N$, and stride $1$.
      Let $M$ be a stacked convolutional layers with linear activation plus a skip connection, i.e., $M(X) = \left( M^{(1)}\circ M^{(2)} \circ \dots \circ M^{(d)} \right) (X) + X$.
      Then, we can choose the right singular vectors of $M$ so that all of them can be represented by $a \otimes (F_{N})_i\otimes(F_{N})_j$ for some $i, j \in\{0,\dots,N-1\}$ and $a \in \mathbb{C}^{m^{(1)}}$.
    \end{prop}

    \begin{prop}
      \label{prop:normalized_conv}
      A convolutional layer followed by a normalization layer such as batch-normalization or weight-normalization
      can be rerepresented as another convolutional layer without normalization at test time.
      Thus, Props.\,\ref{prop:stacked_conv} and \ref{prop:skip_conv} also hold when normalization layers exist.
    \end{prop}

    These propositions show that manipulating a single Fourier basis function on inputs can be most effective to disturb internal representations of convolutional neural networks.

  \subsection{Reduction layers}

    In this section, we show that
    the singular values of the convolutional layers can be written by a combination of a few Fourier basis functions
    even when there are reduction layers, such as convolutional layers with stride $> 1$ or average pooling layers.

    \begin{prop}
      \label{prop:reduction}
      Let $M$ be a convolutional layer with stride $s > 1$ where $N=0$ $(\mathrm{mod}$ $s)$.
      Then, we can choose the right singular vectors of the layer so that all of them can be represented by a linear combination of Fourier basis functions
      $\{(F_N)_{i'}\otimes (F_N)_{j'}|i'=i\ (\mathrm{mod}\ N/s), j'=j\ (\mathrm{mod}\ N/s)\}$ for some $i$ and $j$.
    \end{prop}

    Since the average pooling layer is a special case of convolutional layers, we can apply the above theorem to the layer.    

  \subsection{Single Fourier attack}

    We propose an algorithm to find universal adversarial perturbations using Fourier basis functions.
    The attack exploits the sensitivity of convolutional networks to the Fourier basis directions analyzed in the previous section.
    While the linear approximation in the analysis might not hold well in deep networks, we can still expect that the directions will disturb hidden representations.

    A sketch of the algorithm is as follows.
    We select one Fourier basis function and use it as a UAP.
    The method to select the frequency is described later in this section.
    The sketch of the algorithm is incompatible with the restriction that the inputs must be real.
    To satisfy the condition, we have the following proposition.

    \begin{prop}
      \label{prop:complex}
      $S(x)_{i,j} = S(x)^{*}_{N-i,N-j}$ iff the input $x$ is real-valued, where $S(x)^{*}$ is a conjugate of $S(x)$.
    \end{prop}

\begin{figure}[t]
     \begin{minipage}{.49\hsize}
       \center
       \includegraphics[width=\columnwidth]{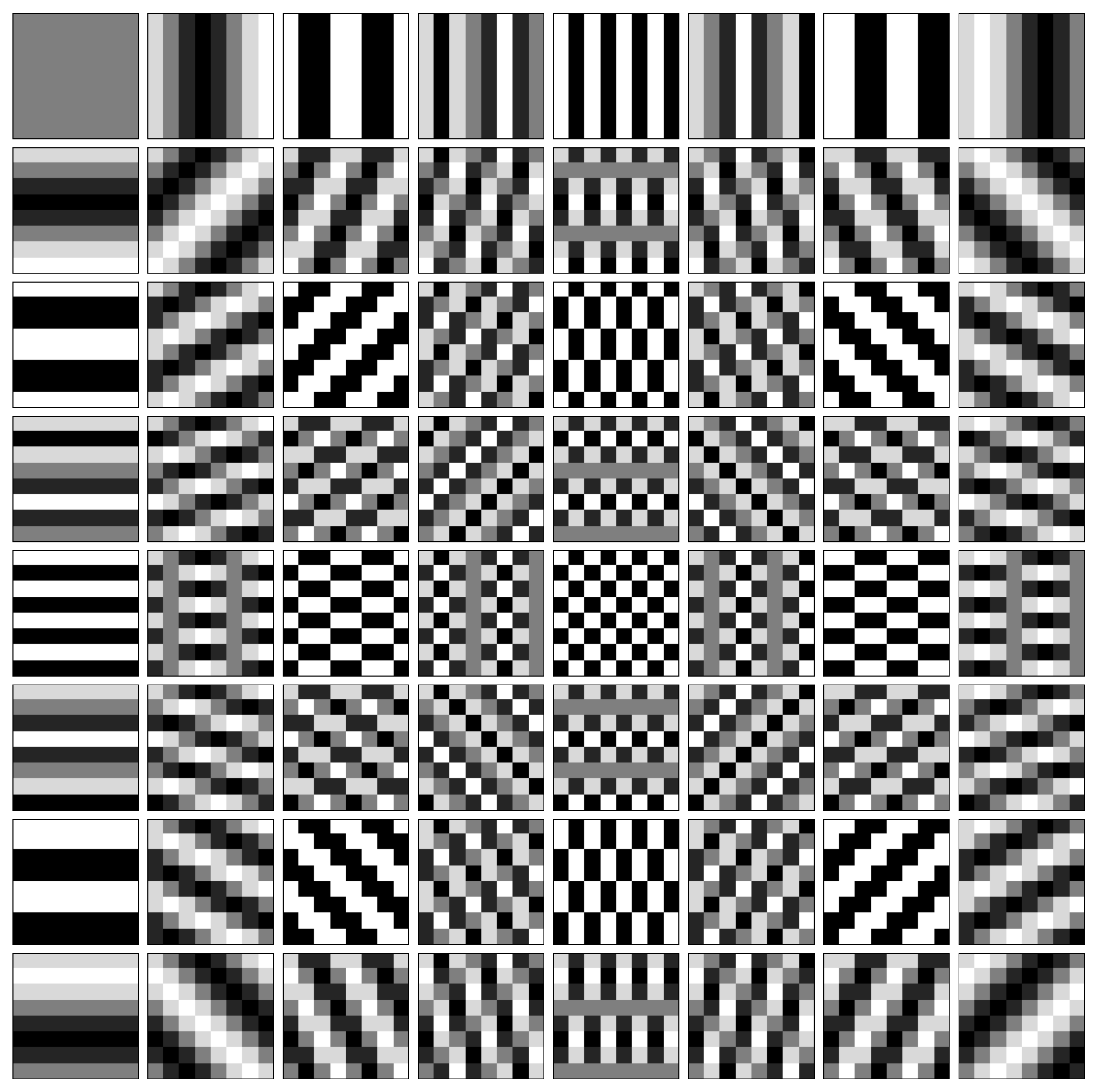}
     \end{minipage}
     \begin{minipage}{.49\hsize}
       \center
       \includegraphics[width=\columnwidth]{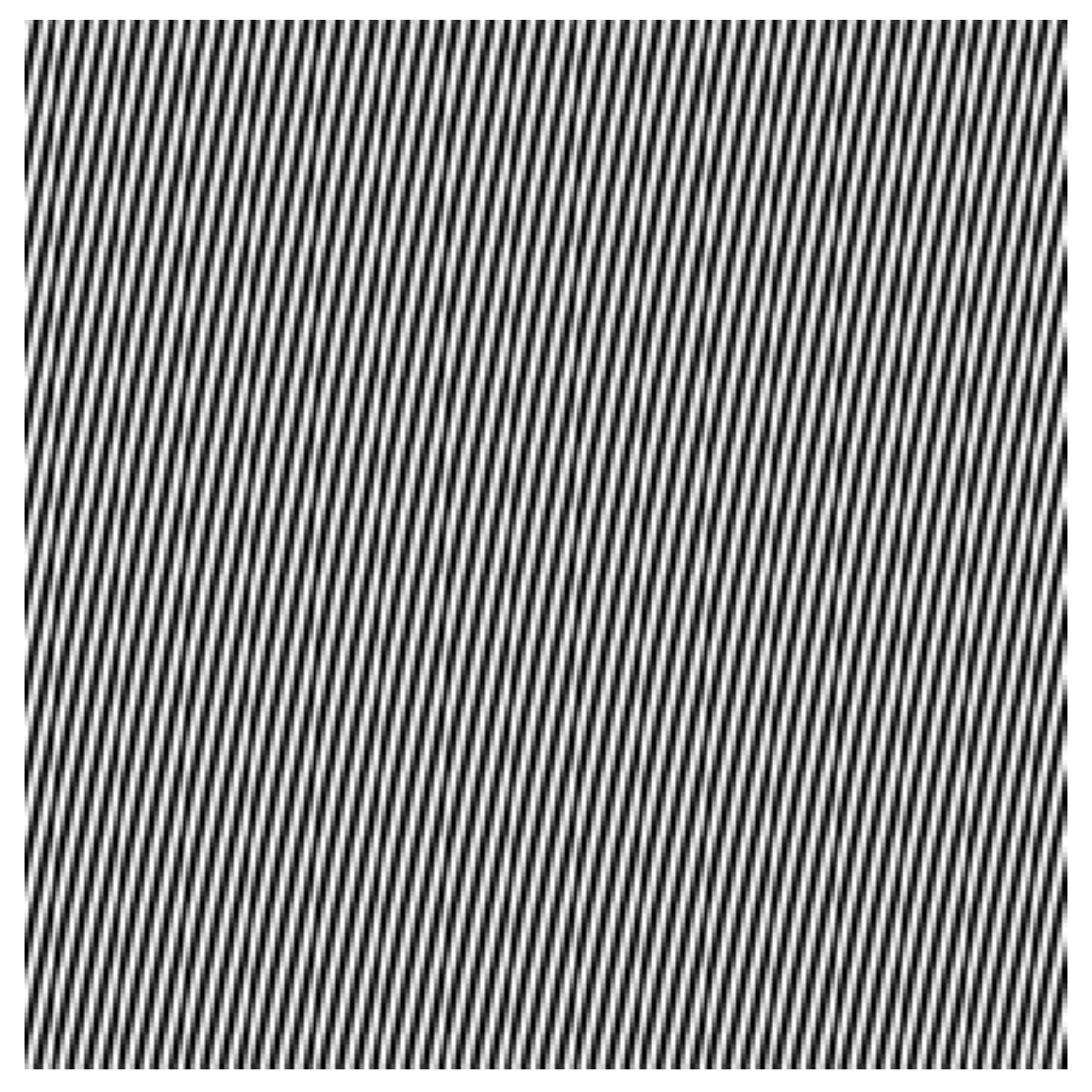}
     \end{minipage}
     \caption{
       Left: Visualization of Fourier basis in $8\times 8$ space.
       Row $i$ and column $j$ shows $(F_8)_i\otimes(F_8)_j$.
       Right:
       An example of perturbations created by Single Fourier attack in Alg.\,\ref{alg:sfa}.
       This perturbation was used in later evaluation\,(\Sec{subsec:fool_ratio}).
     }
     \label{fig:hffm_noise_examples}
   \end{figure}

    Thus, we make $S(x)_{i,j} = S(x)^{*}_{N-i,N-j}$ satisfied to meet the real-value constraint.
    Algorithm\,\ref{alg:sfa} shows the pseudocode of the algorithm, which is named single Fourier attack (SFA).
    Figure\,\ref{fig:hffm_noise_examples} shows a visualization of Fourier basis in $8\times 8$ space and an example of perturbations created by SFA.
    Figure\,\ref{fig:hffm_examples} shows examples of perturbed images.
    It seems that this attack does not change human's predictions, and models should be robust against the attack.

    To perform the attack, we need to find effective frequencies of the target classifiers.
    To test the sensitivity, first, we query a pair of an original image and its perturbed version.
    Next, we check whether the classifier's output differs or not.
    We repeat the procedure and solve a black-box optimization problem formulated as follows.
    \begin{probrem}
      \label{prob:opt}
      Given a data distribution $D \subset \mathbb{R}^{N\times N\times 3}$,
      target function $f: \mathbb{R}^{N\times N\times 3}\times\left(\{1,\dots,N\}\times\{1,\dots,N\}\right) \rightarrow \{0,1\}$,
      find a frequency $w \in \{1,\dots,N\}\times\{1,\dots,N\}$ which maximizes
      \begin{equation}
        \int_D f(x, w) dx.
      \end{equation}
    \end{probrem}

    One naive approach to approximately solve the problem is testing all frequencies with a batch of images and find a frequency with the highest fool ratio.
    The batchsize controls the variance of the evaluation of each frequency.
    Even if we do the brute-force search, we can create UAPs within a reasonable amount of time thanks to the simplicity of our formulation.
    As more query efficient methods, we can also use Bayesian optimization techniques\,\citep{GPUCB,GPEI}.
    We show that the search of the frequency has a favorable property for such methods in Sec.\,\ref{subsec:coocurance}.
    This suggests our algorithm is useful even when only small numbers of queries are allowed to create UAPs.

    Our formulation and algorithm have the following two key benefits.
    First, we formulated the creation of UAPs as an optimization problem of two discrete variables.
    On the other hand, the original problem has the same number of parameters with the input size, which can be tens of thousands.
    This reduction of parameters to optimize is a significant simplification.
    Second, our algorithm requires neither model parameters nor output logits.
    Prior UAPs creation algorithms require access to models or substituted models created by attackers.
    These requirements have made the attacks less practical.
    In our algorithm, we only require the information on the predicted label by the target.
    Thus, the algorithm is available in broader settings.

    \begin{algorithm}[t]
      \SetStartEndCondition{}{}{}
      \SetKwFor{ForEach}{foreach}{:}{fintq}
      \AlgoDontDisplayBlockMarkers\SetAlgoNoEnd\SetAlgoNoLine
      \SetKwFunction{Clip}{Clip}
      \SetKwInOut{hyperparam}{hyperparam}
      \SetKwInOut{input}{input}
      \hyperparam{$i, j$: frequency, $\epsilon$: size of perturbation}
      \input{$x :$ image}
      \ForEach{ c in channel}{
        $x_c \leftarrow x_c + \epsilon ((1 + i)(F_N)_i\otimes(F_N)_j
        $\newline$
        + (1-i)(F_N)_{N-i}\otimes(F_N)_{N-j})$\;
        $x_c \leftarrow $Clip($x_c, 0, 1$)\;
      }
      \caption{Single Fourier attack}
      \label{alg:sfa}
   \end{algorithm}

\section{Experiments}
\label{sec:experiment}
  
  We presented a characterization of the universal adversarial directions through Fourier basis functions in Sec.\,\ref{sec:fourier_analysis}.
  To show that the characterization well describes the nature of the universal adversarial directions, we conducted a series of experiments.
  Primarily, we answer the following questions.
  \begin{enumerate}
    \item Whether Fourier basis characterization is better than others such as characterization using the standard basis (Sec.\,\ref{subsec:fourier_coordinate}).
    \item Whether the sensitivity to the Fourier basis directions is unique to convolutional networks (Sec.\,\ref{subsec:conv_mlp}).
    \item Whether UAPs are related to Fourier basis directions (Sec.\,\ref{subsec:uap_in_fourier}).
    \item Whether current white-box attacks are also related to Fourier basis directions (Sec\,\ref{subsec:fgsm_in_fourier}).
    \item Whether manipulation on a single Fourier basis can image-agnostically change predictions of various convolutional neural networks and datasets (Sec.\,\ref{subsec:fool_ratio}).
  \end{enumerate}

  \subsection{Evaluation setups}

    This section describes the evaluation setups.
    A more detailed explanation can be found in the supplementary material.
    We used MNIST\,\citep{MNIST}, fashion-MNIST\,\citep{FMNIST}, SVHN\,\citep{SVHN}, CIFAR10, CIFAR100\,\citep{CIFAR}, and ILSVRC2015\,\citep{ILSVRC} as datasets.
    We used a multi-layer perceptron (MLP) consisting of $1000$--$1000$ hidden layer with ReLU activation,
    LeNet\,\citep{LeNet}, WideResNet\,\citep{WideResNet}, DenseNet-BC\,\citep{DenseNet}, and VGG\,\citep{VGG} with batch-normalization for evaluations on datasets except for ILSVRC2015.
    For ILSVRC2015, we used ResNet50\,\citep{ResNet}, DenseNet, VGG16, and GoogLeNet\,\citep{GoogLeNet}.
    For VGG16 and GoogLeNet, we added a batch-normalization layer after each convolution for faster training.
    We used the fool ratio as a metric, which is the percentage of data that models changed their predictions, following \citet{Moosavi{-}Dezfooli \etal}{Universal}.

  \subsection{Fourier domain vs pixel domain}
    \label{subsec:fourier_coordinate}

\begin{figure}[t]
      \begin{minipage}{\hsize}
          \center
          \includegraphics[width=\columnwidth]{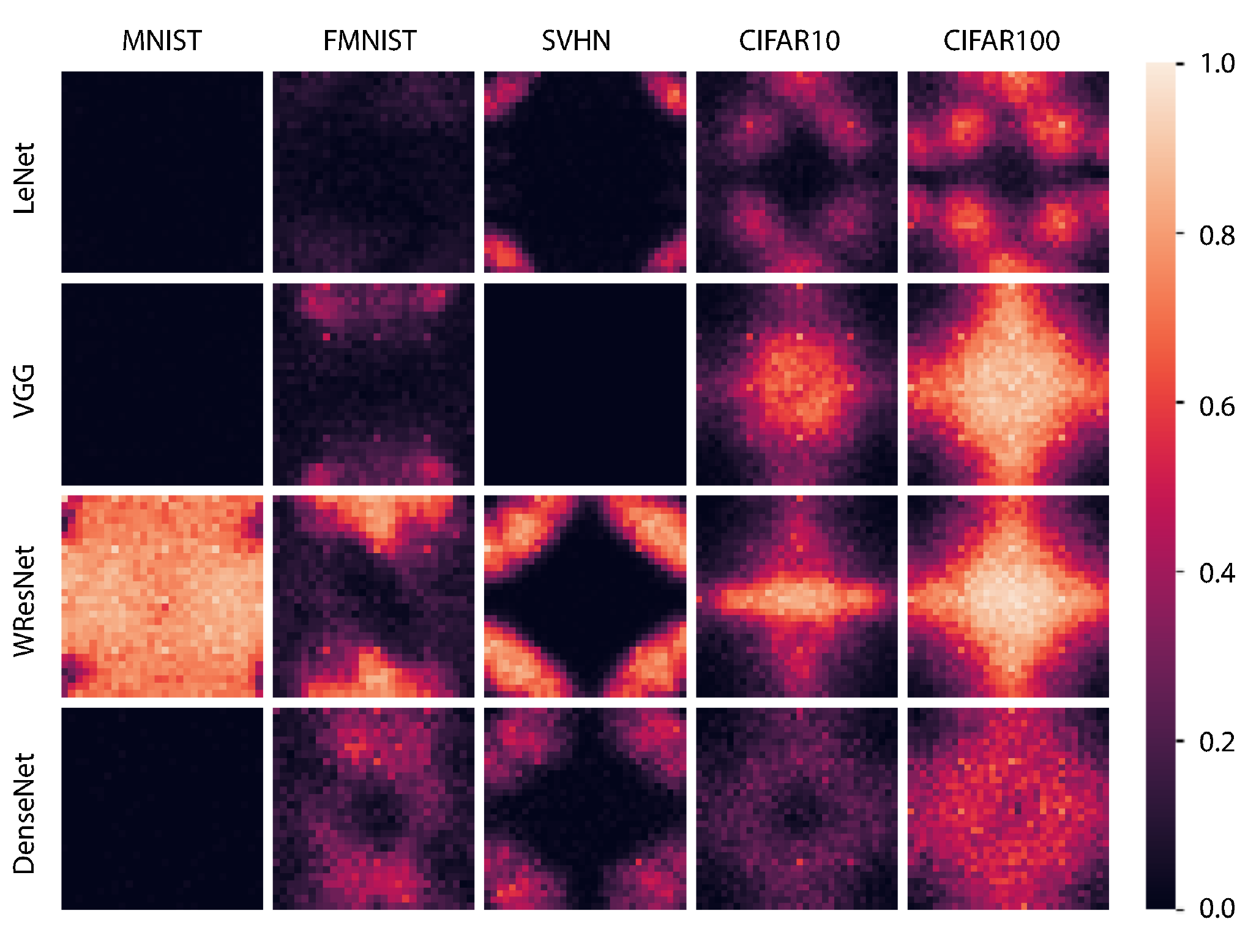}
      \end{minipage}
      \caption{
        Visualization of sensitive spot of convolutional networks in Fourier domain.
        Coordinate $(i, j)$ of each image represents fool ratio on a single minibatch when we used Algorithm\,\ref{alg:sfa} as a perturbation.
        White areas are spots with high fool ratio.
        The center of each image corresponds to a high-frequency area.
        The perturbation sizes were $30/255$ for MNIST, FMNIST, and SVHN, and $10/255$ for CIFAR10 and CIFAR100.
        The creation of this heatmap is described in Algorithm.\,\ref{alg:heatmap}.
      }
      \label{fig:fourier_coordinate}
    \end{figure}

\begin{figure}[t]
      \begin{minipage}{\hsize}
        \center
        \includegraphics[width=\columnwidth]{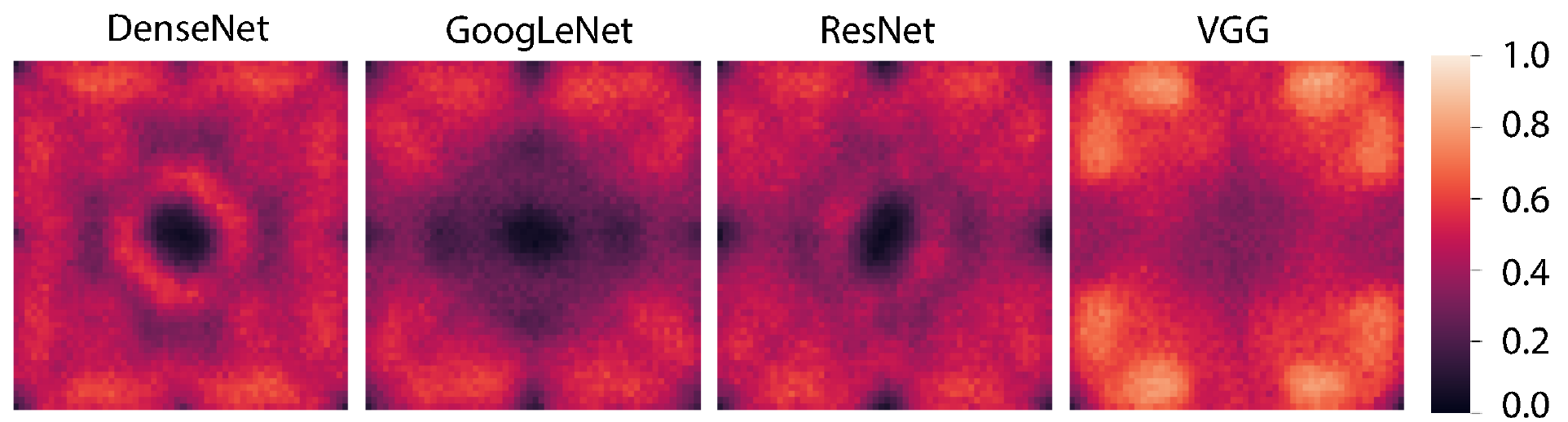}
      \end{minipage}
      \caption{
        Visualization of sensitivity in Fourier domain.
        Visulaization procedure is the same with Figure\,\ref{fig:fourier_coordinate}.
        We can see that most sensitive frequency is neither hight nor low frequencies, and it lies in the middle.
        For reference, frequency distributions in natural images and random noise can be found in Figure\,\ref{fig:uap_in_fourier}.
      }
      \label{fig:fourier_coordinate_ilsvrc}
    \end{figure}

    We analyzed the sensitivity of deep convolutional neural networks to the directions of Fourier basis functions in Sec.\,\ref{sec:fourier_analysis}.
    To empirically support the analysis, we investigated the sensitivity on each Fourier basis function.
    For comparison, we checked the sensitivity on the standard basis directions, which is the manipulation on each pixel.
    We also tested the sensitivity in random directions (see Sec.\,\ref{subsec:fool_ratio}).
    We first describe the method we used to study the sensitivity.
    For Fourier basis, we applied a single Fourier attack (Algorithm.\,\ref{alg:sfa}) and calculated its fool ratio on a single minibatch for each frequency.
    We bounded the size of perturbations by $30/255$ in $\ell_\infty$-norm for MNIST, FMNIST, and SVHN, $20/255$ for ILSVRC2015, and $10/255$ for CIFAR10 and CIFAR100.
    For a standard basis, we added $255/255$ to each pixel and then clipped to range from zero to one for attack creation, which is an analogy of Algorithm\,\ref{alg:sfa}.
    Using heat maps, we visualized the results for Fourier basis on ILSVRC2015 in Figure\,\ref{fig:fourier_coordinate_ilsvrc} and the results on the other datasets in Figure\,\ref{fig:fourier_coordinate}.
    The algorithm to create the heat map is described in Algorithm\,\ref{alg:heatmap}.
    \begin{algorithm}[t]
      \SetStartEndCondition{}{}{}
      \SetKwFor{ForEach}{foreach}{:}{fintq}
      \AlgoDontDisplayBlockMarkers\SetAlgoNoEnd\SetAlgoNoLine
      \SetKwFunction{Forward}{Forward}
      \SetKwFunction{FoolRatio}{FoolRatio}
      \ForEach{ (i,j) in frequencies}{
        $B :=$ Randomly select Minibatch\;
        $y \leftarrow $ Forward($B$)\;
        $y' \leftarrow $ Forward($B$ + noise)\;
        Heatmap$_{i,j}$ $\leftarrow$ FoolRatio($y,y'$)\;
      }
      \caption{Creation of heatmap}
      \label{alg:heatmap}
   \end{algorithm}

    We observed that in most cases except for MNIST, architectures tend to have some sensitive spots in the Fourier domain.
    Especially on CIFAR10 and CIFAR100, VGG and Wide-ResNet showed near $90\%$ and $99\%$ fooling ratio to some directions.
    The result means that the predictions became almost random guess.
    Since all Fourier basis directions are orthogonal, Figure\,\ref{fig:fourier_coordinate} highlights that there are hundreds of directions
    that networks are sensitive independent of their inputs.
    While it has been known that there are tens of orthogonal directions for transferable or universal adversarial examples,
    to the best of our knowledge, this is the fastest method to find a large number of orthogonal directions for which networks are universally vulnerable.
    Contrastive to Fourier basis, experiments using standard basis achieved almost $0\%$ fool ratio in all settings.
    In this experiment, we showed the existence of sensitive spots of convolutional networks in the Fourier domain and the effectiveness of the characterization by Fourier basis directions.

  \subsection{Convolutional networks vs. MLP}
    \label{subsec:conv_mlp}
\begin{figure}[t]
      \begin{minipage}{\hsize}
        \center
        \includegraphics[width=\columnwidth]{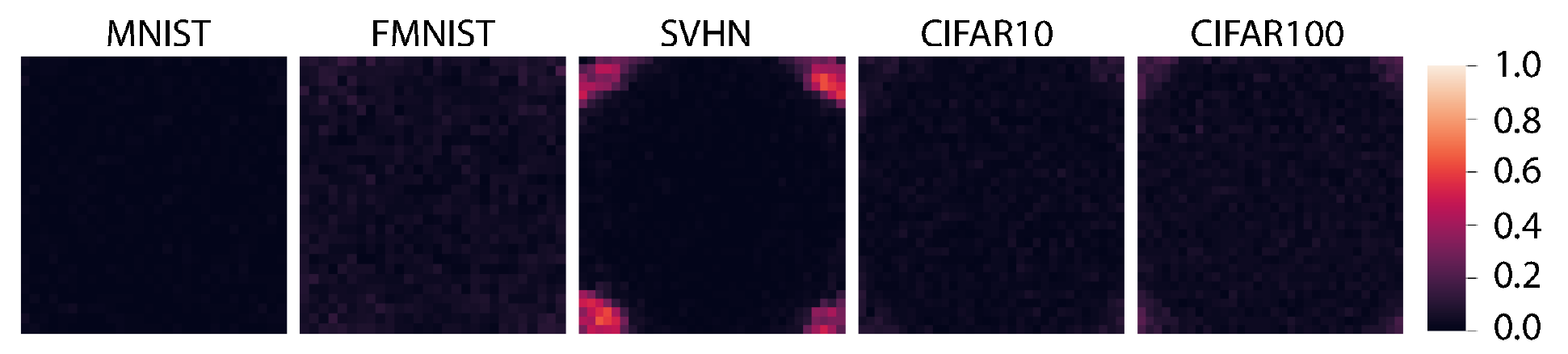}
      \end{minipage}
      \caption{
        Visualization of the sensitivity of multilayer perceptrons (MLPs) in the Fourier domain.
        MLPs did not have sensitive spot as CNNs in most cases and they were more resistant to directions of Fourier basis.
      }
      \label{fig:cifar_compare}
    \end{figure}

    In Sec.\,\ref{subsec:fourier_coordinate} we observed that various convolutional neural networks are sensitive to some Fourier basis directions.
    To see whether the sensitivity to the Fourier basis functions is caused by network architectures as suggested in Sec.\,\ref{sec:fourier_analysis} or the nature of image processing, we compared the sensitivity of a MLP to Fourier basis functions.
    We used the same method as Sec.\,\ref{subsec:fourier_coordinate} for the comparison.
    Figure\,\ref{fig:cifar_compare} shows the results for the MLP trained on various datasets.
    The MLP did not show the vulnerability to some vectors in the Fourier basis.
    The contrastive activation pattern of convolutional networks and multilayer perceptrons supports our analysis of the sensitivity in Sec.\,\ref{sec:fourier_analysis}.
    This result suggests the possibility that changing architectures is a useful measure to mitigate adversarial examples, especially UAPs.
    Since prior defense work has mostly focused on training methods\,\citep{FGSM,AtScale}, this opens another research direction for defense methods.
    For example, we may use the information of the weak spots in the Fourier domain to choose which models to use for ensembles.

  \subsection{Co-occurrence of sensitivity}
    \label{subsec:coocurance}
\begin{figure}[t]
      \begin{minipage}{\hsize}
        \begin{minipage}{.24\hsize}
          \center
          \includegraphics[width=\columnwidth]{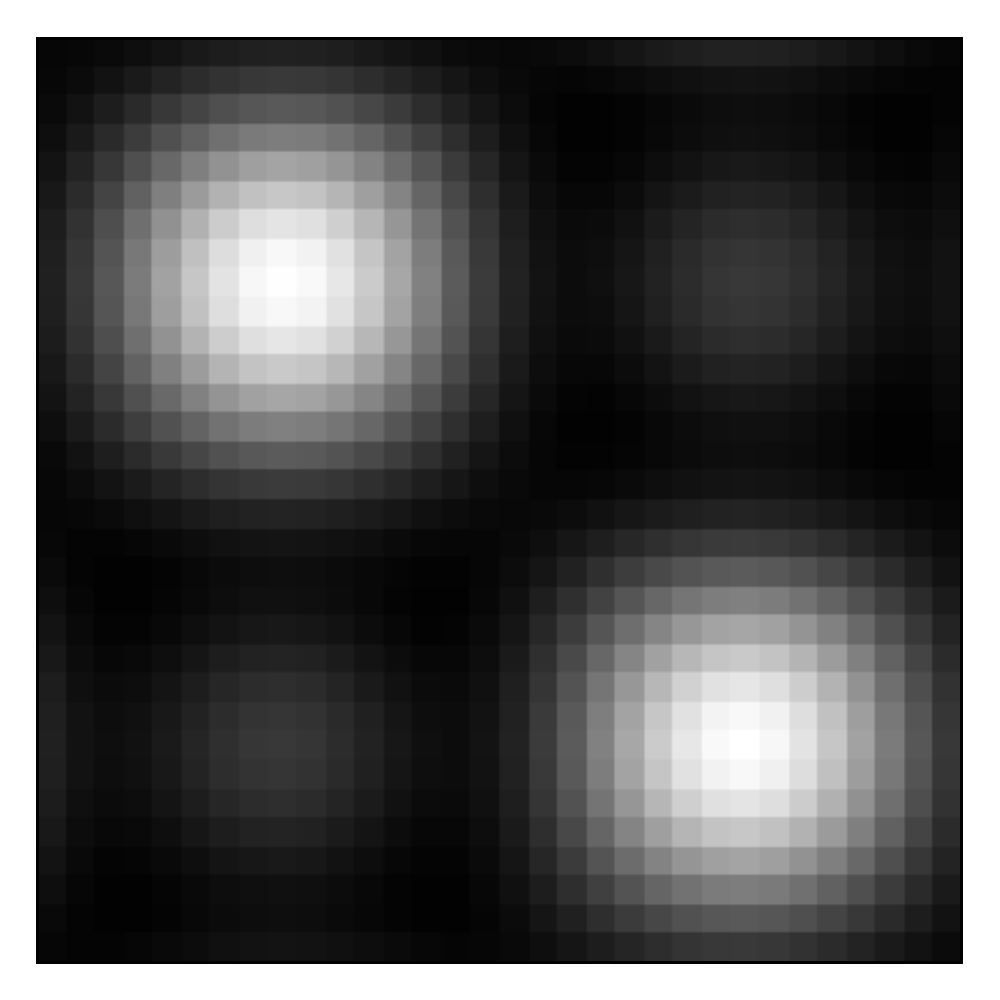}
        \end{minipage}
        \begin{minipage}{.24\hsize}
          \center
          \includegraphics[width=\columnwidth]{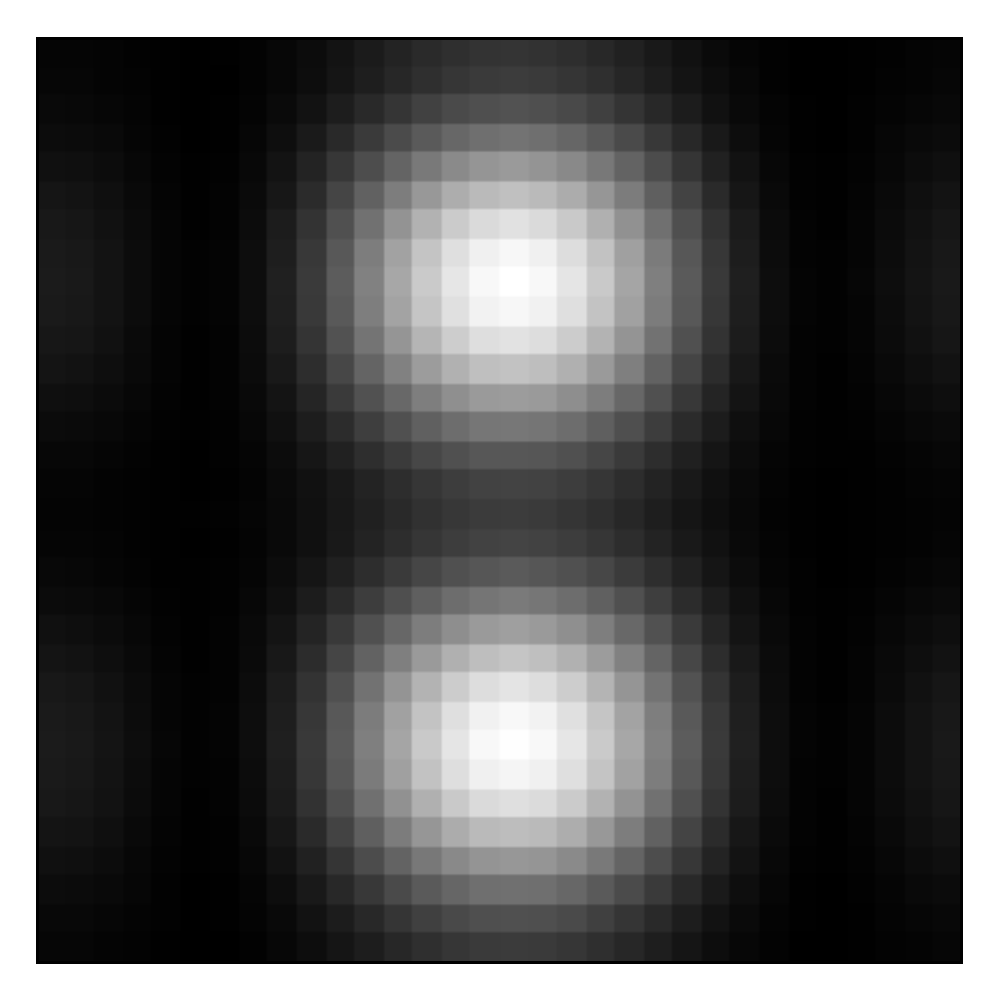}
        \end{minipage}
        \begin{minipage}{.24\hsize}
          \center
          \includegraphics[width=\columnwidth]{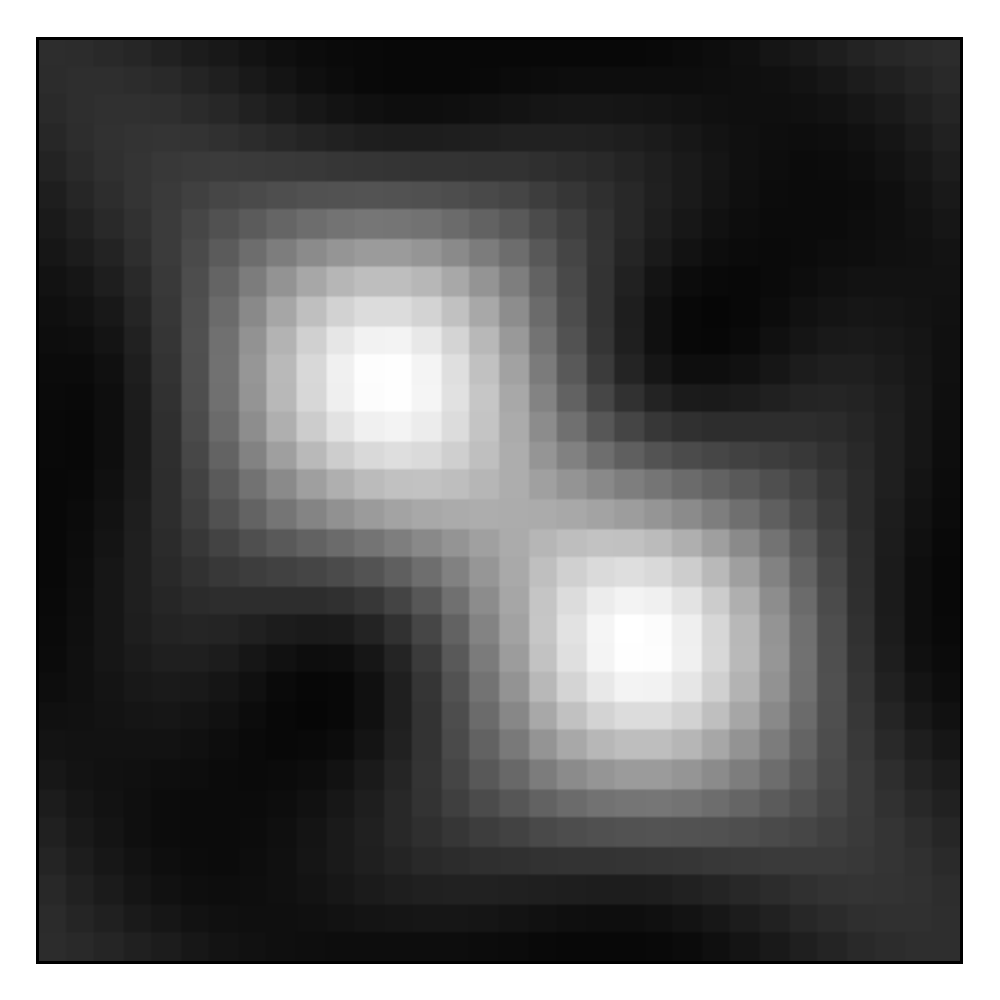}
        \end{minipage}
        \begin{minipage}{.24\hsize}
          \center
          \includegraphics[width=\columnwidth]{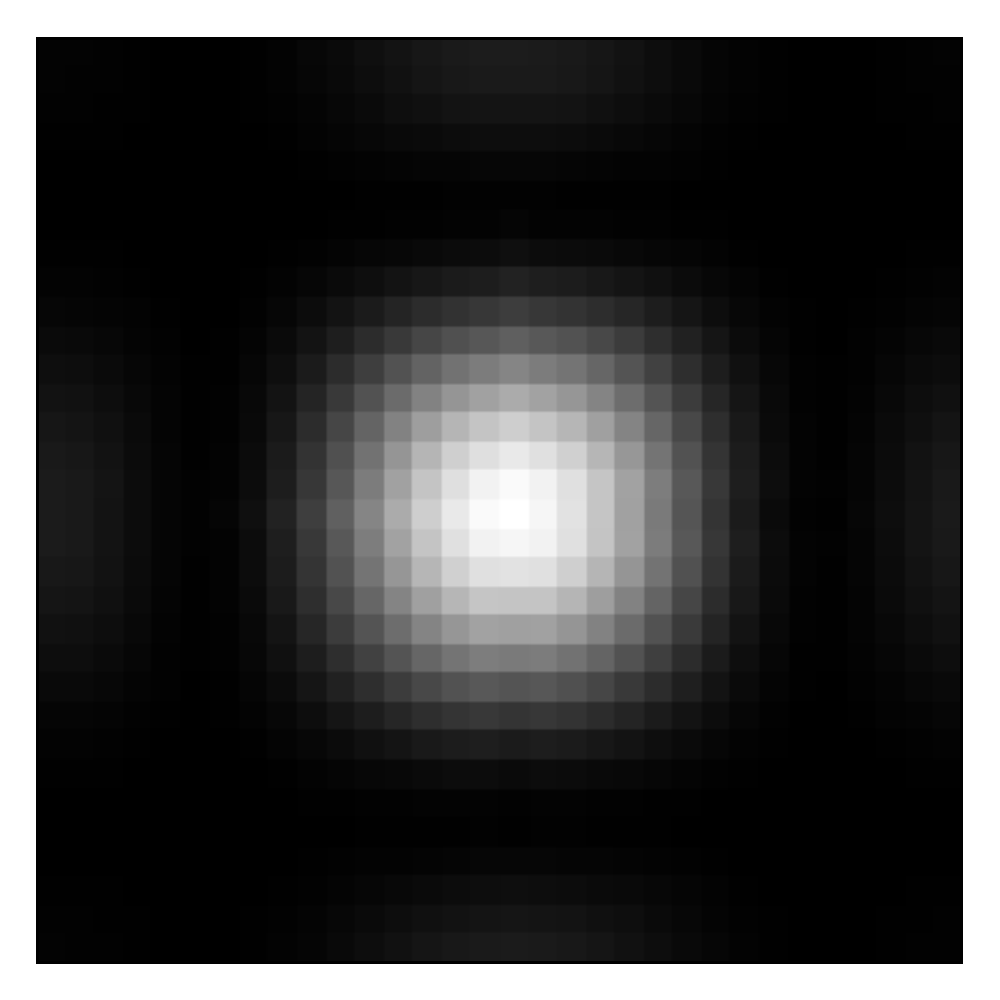}
        \end{minipage}
      \end{minipage}
      \caption{
        Coordinate $(i, j)$ of each image shows the magnitude of outputs of a convolutional layer
        when input was $(F_{32})_{i}\otimes(F_{32})_{j}$.
        The kernel size of each convolutional layer is $3$.
        They were trained to maximize the output against
        $(F_{32})_{8}\otimes(F_{32})_{8}, (F_{32})_{8}\otimes(F_{32})_{16}, (F_{32})_{12}\otimes(F_{32})_{12}, and (F_{32})_{16}\otimes(F_{32})_{16}$, respectively.
      }
      \label{fig:coocurance}
    \end{figure}

    In the evaluation in Secs.\,\ref{subsec:fourier_coordinate} and \ref{subsec:conv_mlp},
    we observed that convolutional networks showed similar sensitivity to the Fourier basis directions with similar frequencies.
    Since Sec.\,\ref{sec:fourier_analysis} does not cover this phenomenon, we explain it here.
    In convolutional networks, the convolution kernel size is typically much smaller than the input size.
    The size of the kernel restricts the expressiveness of convolutional layers.
    This restriction makes convolutional layers respond similarly to similar frequencies.
    To see the co-occurrence of the sensitivity, we trained convolutional layers with kernel size $3\times3$ and the input size $32\times32$ so that the $\ell_2$-norm of their outputs are maximized
    when one specific Fourier basis is fed as its input.
    Then we tested the $\ell_2$-norm of the layer's outputs when their inputs are other Fourier basis functions.
    Figure\,\ref{fig:coocurance} shows the result.
    The result confirms the hypothesis that convolutional layers respond similarly to Fourier basis directions with similar frequency.
    In other words, the optimization problem\,\ref{prob:opt} has a small Lipschitz constant.
    This property is known to be favorable for optimizations in many algorithms including Bayesian optimizations\,\citep{GPUCB}.

  \subsection{UAPs in Fourier domain}
    \label{subsec:uap_in_fourier}
\begin{figure}[t]
      \includegraphics[width=\columnwidth]{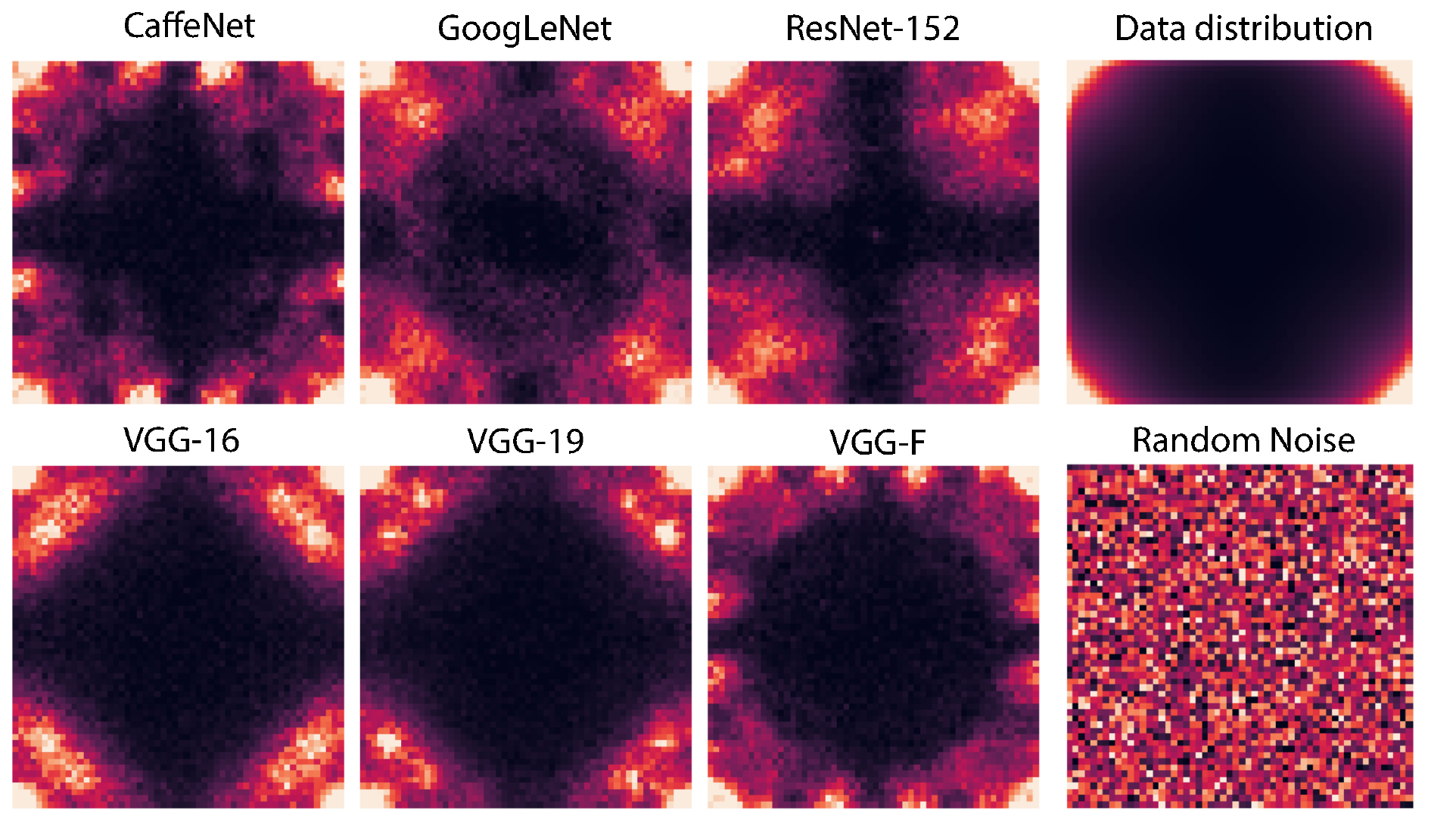}
      \caption{
        Visualization of UAPs calculated for various architectures on ILSVRC2012 by \citet{Moosavi{-}Dezfooli \etal}{Universal} in the Fourier domain.
        Coordinate $(i, j)$ corresponds to the fool ratio of $(F_{224})_{i}\otimes(F_{224})_{j}$.
        White spots had higher fool ratio.
      }
      \label{fig:uap_in_fourier}
    \end{figure}

    In this section, we investigate whether UAPs created by an existing method also have some specific patterns in the Fourier domain.
    For this analysis, we used precomputed UAPs for VGG16, VGG19, VGG-F, CaffeNet\,\citep{CaffeNet}, ResNet152, and GoogLeNet by \citet{Moosavi{-}Dezfooli \etal}{Universal}.
    Figure\,\ref{fig:uap_in_fourier} shows the magnitude of each frequency of each UAP in log scale.
    For reference, Figure\,\ref{fig:uap_in_fourier} also shows those of random noise and average magnitudes of each frequency of original training data in ILSVRC2015.

    While architectures and training procedures differ, Figure\,\ref{fig:uap_in_fourier} and Figure\,\ref{fig:fourier_coordinate_ilsvrc} share a similar tendency compared to the random noise and original images.
    For example, we can see from Figure\,\ref{fig:fourier_coordinate_ilsvrc} that the networks are relatively robust against high-frequency noises and sensitive to low and middle-frequency noises.
    From Figure\,\ref{fig:uap_in_fourier}, current UAPs appear to exploit the sensitivity.
    This suggests the effectiveness to consider Fourier domain to analyze existing UAPs.

  \subsection{Adversarial attacks in Fourier domain}
    \label{subsec:fgsm_in_fourier}
\begin{figure}[t]
      \begin{minipage}{\hsize}
        \center
        \includegraphics[width=\columnwidth]{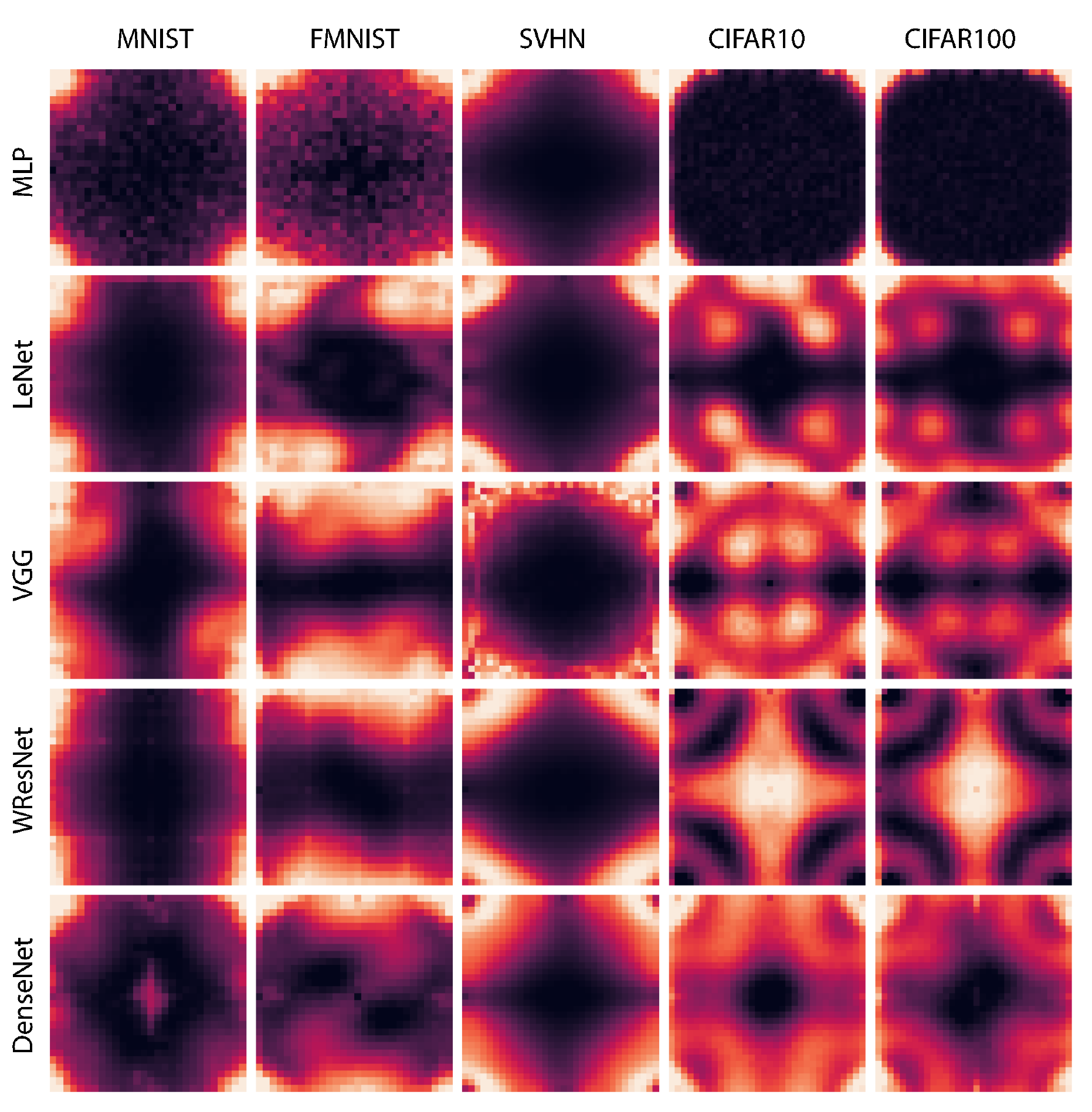}
      \end{minipage}
      \caption{
        Visualization of FGSM attack in the Fourier domain in the same way as in Figure\,\ref{fig:uap_in_fourier}.
        FGSM had larger values in sensitive spots revealed in Figure\,\ref{fig:fourier_coordinate}.
        Center of each image is a high-frequency area.
      }
      \label{fig:fgsm_in_fourier}
    \end{figure}

    In this section, we investigate whether current white box adversarial attacks also have some tendency in the Fourier domain.
    We studied FGSM\,\citep{FGSM}, which is known to transfer better than naive iterative attacks\,\citep{AtScale}.
    Figure\,\ref{fig:fgsm_in_fourier} shows the average magnitude of each vector in the Fourier basis of a perturbation created by FGSM on test data.
    Compared with Figure\,\ref{fig:fourier_coordinate}, which revealed sensitive spots in the Fourier domain, Figure\,\ref{fig:fgsm_in_fourier} shows that the mass of FGSM concentrates almost in the sensitive spots.
    This experiment also shows that adversarial perturbations do not necessarily lie in a high-frequency area, which denies a common myth that adversarial perturbations tend to be high-frequency.
    Figure\,\ref{fig:fgsm_in_fourier} also shows that the tendency of adversarial perturbations differs across datasets and architectures, which reminds us to test defense methods in various settings.

  \subsection{Effectiveness of Fourier attack}
    \label{subsec:fool_ratio}
    The analysis in Sec.\,\ref{sec:fourier_analysis} and experiments in Secs.\,\ref{subsec:fourier_coordinate} -- \ref{subsec:fgsm_in_fourier} suggests the effectiveness of the Fourier basis functions as universal adversarial perturbations.
    We evaluated its ability to flip predictions on various datasets and architectures.
    We set the size of perturbations to $10/255$ in $\ell_\infty$ for CIFAR, and to $30/255$ for MNIST, FMNIST, and SVHN.
    We used frequencies with the highest fool ratio in Figure\,\ref{fig:fourier_coordinate} as the perturbations.
    In the evaluation, we used Algorithm\,\ref{alg:sfa} with one fixed frequency per pair of dataset and architecture.
    For comparison, we calculated the fool ratio of random noise sampled from the $\epsilon$-ball bounded in $\ell_\infty$-norm.
    Table\,\ref{table:fool_ratio} shows the result.
    \begin{table}[t]
        \caption{
          Fool ratio of random noise (upper rows) and SFA (Algorithm\,\ref{alg:sfa}, lower rows) on various architectures and datasets.
          Despite the simpleness of our algorithm, some pairs dropped their accuracy to almost chance ratio.
          This results show that our characterization through Fourier basis functions effectively captures the sensitivity of networks.
        }
        \label{table:fool_ratio}
        \small
        \begin{center}
            \begin{tabular}{ c r r r r }
                \toprule
                              & LeNet  & WResNet & VGG    & DenseNet \\
                \midrule
                MNIST         & $ 0.1$ & $55.8$  & $ 0.0$ & $ 0.1$ \\
                Fashion MNIST & $ 5.4$ & $11.4$  & $ 8.3$ & $12.6$ \\
                SVHN          & $ 3.1$ & $ 5.2$  & $0.0$  & $ 4.6$ \\
                CIFAR10       & $ 5.0$ & $ 8.1$  & $6.8$  & $ 5.4$ \\
                CIFAR100      & $13.0$ & $26.4$  & $25.9$ & $22.5$ \\
                \midrule
                MNIST         & $ 0.4$          & $\mathbf{90.2}$  & $0.1$           & $0.2$ \\
                Fashion MNIST & $12.5$          & $\mathbf{48.1}$  & $\mathbf{83.7}$ & $\mathbf{56.9}$ \\
                SVHN          & $\mathbf{64.9}$ & $\mathbf{90.8}$  & $0.0$           & $\mathbf{50.5}$ \\
                CIFAR10       & $\mathbf{63.3}$ & $\mathbf{82.3}$  & $\mathbf{72.2}$ & $\mathbf{50.7}$ \\
                CIFAR100      & $\mathbf{83.4}$ & $\mathbf{93.7}$  & $\mathbf{95.8}$ & $\mathbf{72.3}$ \\
                \bottomrule
            \end{tabular}
        \end{center}
    \end{table}
    \begin{table}[t]
        \caption{
          Fool ratio of Fourier basis attack on various architectures on ILSVRC2015.
          Rand is random noise, SSFA is defined in Sec.\,\ref{subsec:fool_ratio}.
          Attacks are bounded in $10/255$ and $20/255$ in $\ell_\infty$-norm.
          UAP denotes the best performing precomputed UAP in \citep{Universal} per architecture.
          Since naively scaling them to $20/255$ can be unfairly advantageous to ours and we just omitted the evaluation.
          While comparable, our algorithm does not assume access to the same training data and also has no need to train models locally.
        }
        \label{table:fool_ratio_ilsvrc}
        \small
        \begin{center}
            \begin{tabular}{ c r r r r }
                \toprule
                        & GoogLeNet & ResNet & VGG  & DenseNet \\
                \midrule
                Rand($10$) & $8.2$     & $8.5$    & $11.5$ & $9.5$ \\
                Rand($20$) & $14.9$    & $16.1$   & $19.7$ & $16.7$ \\
                \midrule
                UAP($10$)\,\citep{Universal}  & $45.5$    & $49.7$   & $64.8$ & $56.0$ \\
                \midrule
                SFA($10$)  & $34.6$  & $38.7$   & $49.7$ & $36.8$ \\
                SFA($20$)  & $62.3$  & $68.5$   & $76.3$ & $63.5$ \\
                \midrule
                SSFA($10$) & $44.1$ & $40.1$   & $53.3$ & $39.5$ \\
                SSFA($20$) & $74.1$ & $66.9$   & $79.0$ & $62.5$ \\
                \bottomrule
            \end{tabular}
        \end{center}
    \end{table}
    Given the dataset and architecture-agnostic search space, the attack showed strong attack ability.
    Especially in CIFAR10 and CIFAR100 experiments, some architectures dropped prediction accuracy almost to that of random guessing.
    This effectiveness of Fourier basis attack highlights the sensitivity of current convolutional
    networks against Fourier features.
    In MNIST, however, the fool ratio was not as high as other datasets.
    Since MNIST is highly normalized dataset and easiest among them, we suspect that networks can better capture true signal from the inputs and are more robust to change of a single Fourier basis direction.
    From the viewpoints of architectures, LeNet and DenseNet were more robust than others.
    We explain this by their max-pooling layers.
    As max-pooling layers are not supported in Sec.\,\ref{sec:fourier_analysis}, they add additional nonlinearities and mix Fourier basis.

    We also tested Algorithm\,\ref{alg:sfa} on ILSVRC2015.
    For the evaluation, we fixed one frequency for all architectures and inputs\footnote{The input sizes were the same $(224\times224)$ among the all architectures we tested.}.
    In other words, we selected a single perturbation input and architecture agnostically.
    To choose the frequency, we took the average of Figure\,\ref{fig:fourier_coordinate_ilsvrc} and picked the frequency with the highest fool ratio.
    Figure\,\ref{fig:hffm_examples} shows examples of created adversarial examples.
    We used $10/255$ and $20/255$ for the size of perturbations.
    Note that previous work used $10/255$ for the evaluation\,\citep{Universal}.
    Examples of created UAPs are shown in Figure\,\ref{fig:hffm_examples}.
    We empirically found that taking the sign of Fourier basis can sometimes boost the performance of the attack.
    We named this attack Signed-SFA (SSFA), and we also tested the attack.
    In the evaluation, we also tested random perturbations and the best precomputed UAP from \citet{Moosavi{-}Dezfooli \etal}{Universal} per architecture.
    The result is shown in Table\,\ref{table:fool_ratio_ilsvrc}.
    Compared to \citet{Moosavi{-}Dezfooli \etal}{Universal}, the fool ratio is comparable to their perturbations under this black-box setting.
    Note, since our algorithm does not need to train local model, our algorithm is more suitable in black-box settings.

\section{Conclusion}

  From the analysis of linearized convolutional neural networks, we hypothesized that convolutional networks are sensitive to the directions of Fourier basis functions.
  Through empirical evaluations, we validated the sensitivity.
  The finding provides a better characterization of universal adversarial perturbations using Fourier basis functions.
  The characterization might be beneficial to the development of defense methods and the analysis of statistical generalization guarantees.
  As a by-product of our analysis, we proposed a black-box method to create universal adversarial perturbations.
  The algorithm does not require locally trained models for black-box attack and extends the potential use cases of universal adversarial perturbations.

\section*{Acknowledgement}

  YT was supported by Toyota/Dwango AI scholarship. IS was supported by KAKENHI
17H04693.

\clearpage

  {\small
    \bibliographystyle{ieee}
    \bibliography{cvpr}
  }

\clearpage

\onecolumn

\appendix

\section{Notations}
\label{sec:notation}

  Notations are summarized in table\,\ref{table:notation}.

  \begin{table}[t]
    \caption{
      Notation table.
    }
    \label{table:notation}
    \small
    \begin{center}
      \begin{tabular}{ l }
        \toprule
          $\Circ(c)$: A Circulant matrix crated by a vector $c$.\\
          $x_{i}$: An $i$-th element of a vector $x$.\\
          $A_{i,j}$: An $i$-th row $j$-th column element of a matrix $A$.\\
          $A_{i,j}$: An $i$-th row $j$-th column element of a matrix $A$.\\
          $\omega_N$: $n$-th root of $1$, $\exp(2\pi\sqrt{-1}/N)$.\\
          $\omega_N^i$: $n$-th root of $1$ power $i$.\\
          $F_N$: A matrix which $(F_N)_{i,j} = \omega_N^{(i+j)}$.\\
          $S(X)$: $2$d Fourier transformation of a matrix $X$.\\
          $Q_N$: A matrix $F_N \otimes F_N$.\\
          $\otimes$: A Kronecker product.\\
          $I_m$: $m$-dimensional identity matrix.\\
          $R$: $I_m \otimes Q_N.$\\
          $m$: Channel size.\\
        \bottomrule
      \end{tabular}
    \end{center}
  \end{table}

  \noindent

\section{Preliminary}

  \subsection{Circulant matrix}

    Let $c$ be a vector and $c_i$ be the $i$-th element of the vector $c$.
    A circulant matrix is a matrix with the following shape.
    \begin{equation}
      \Circ(c_:)={\begin{bmatrix}c_{0}&c_{{1}}&\dots &c_{{n-2}}&c_{{n-1}}\\
      c_{{n-1}}&c_{0}&c_{{1}}&&c_{{n-2}}\\
      \vdots &c_{{n-1}}&c_{0}&\ddots &\vdots \\
      c_{{2}}&&\ddots &\ddots &c_{{1}}\\c_{{1}}&c_{{2}}&\dots &c_{{n-1}}&c_{0}\\\end{bmatrix}}.
    \end{equation}
    A doubly block circulant matrix is a block matrix whose blocks are circulant.
    The matrix $A$ below is an example of a doubly block circulant matrix.
    \begin{equation}
      A={\begin{bmatrix}
      \Circ(K_{0,:})&\Circ(K_{1,:})&\dots &\Circ(K_{n-2,:})&\Circ(K_{n-1,:})\\
      \Circ(K_{n-1,:})&\Circ(K_{0,:})&\Circ(K_{1,:})&&\Circ(K_{n-2,:})\\
      \vdots &\Circ(K_{n-1,:})&\Circ(K_{0,:})&\ddots &\vdots \\
      \Circ(K_{2,:})&&\ddots &\ddots &\Circ(K_{1,:})\\
      \Circ(K_{1,:})&\Circ(K_{2,:})&\dots &\Circ(K_{n-1,:})&\Circ(K_{0,:})\\\end{bmatrix}},
    \end{equation}
    where $K_{i,:}$ is a $i$-th row of a matrix $K$.
    When the channel size of a convolutional layer is equal to one and padding is ``wraps around,''
    convolution operation can be written as a doubly block circulant matrix \citep{DeepLearning,SingularConv}.

\section{Proof of propositions}

  \subsection{Proposition\,\ref{prop:decomposition}}

    We prove the proposition following \citet{Sedghi \etal}{SingularConv}.
    Our assumption is that the padding is ``wrap around''.
    Under the assumption, a convolutional can be represented by the following matrix $M$.
    \begin{equation}
    \label{eq:conv}
      M = {
        \begin{bmatrix}
          B^{(0,0)}&B^{(0,1)}&\dots&B^{(0, m_{\mathrm{in}}-1)}\\
          B^{(1,0)}&B^{(1,1)}&\dots&B^{(1, m_{\mathrm{in}}-1)}\\
          \vdots&\vdots&\ddots&\vdots\\
          B^{(m_{\mathrm{out}} - 1,0)}&B^{(m_{\mathrm{out}} - 1,1)}&\dots&B^{(m_{\mathrm{out}} - 1, m_{\mathrm{in}}-1)}\\
        \end{bmatrix}
      },
    \end{equation}
    where each $B^{(c,d)}$ is a doubly circulant matrix.
    Let $D^{(c,d)} = Q_N^{\mathrm{H}}B^{(c,d)}Q_N$.
    Since $B_{c,d}$ is a doubly circulant matrix, $D^{(c,d)}$ is a diagonal matrix.
    Now we can write,
    \begin{equation}
      \left(I_{\mathrm{out}}\otimes Q_N\right)^{\mathrm{H}}M\left(I_{\mathrm{in}}\otimes Q_N\right) = {
        \begin{bmatrix}
          D^{(0,0)}&D^{(0,1)}&\dots&D^{(0, m_{\mathrm{in}}-1)}\\
          D^{(1,0)}&D^{(1,1)}&\dots&D^{(1, m_{\mathrm{in}}-1)}\\
          \vdots&\vdots&\ddots&\vdots\\
          D^{(m_{\mathrm{out}} - 1,0)}&D^{(m_{\mathrm{out}} - 1,1)}&\dots&D^{(m_{\mathrm{out}} - 1, m_{\mathrm{in}}-1)}\\
        \end{bmatrix}
      }.
    \end{equation}
    By multiplying $\left(I_{m_{\mathrm{out}}}\otimes Q_N\right)$ from left and $\left(I_{m_{\mathrm{in}}}\otimes Q_N\right)^{\mathrm{H}}$ from right, we have
    \begin{equation}
      M = {
        \left(I_{m_{\mathrm{out}}}\otimes Q_N\right) L \left(I_{m_{\mathrm{in}}}\otimes Q_N\right)^{\mathrm{H}}
      },
    \end{equation}
    where
    \begin{equation}
      L = {
        \begin{bmatrix}
          D^{(0,0)}&D^{(0,1)}&\dots&D^{(0, m_{\mathrm{in}}-1)}\\
          D^{(1,0)}&D^{(1,1)}&\dots&D^{(1, m_{\mathrm{in}}-1)}\\
          \vdots&\vdots&\ddots&\vdots\\
          D^{(m_{\mathrm{out}} - 1,0)}&D^{(m_{\mathrm{out}} - 1,1)}&\dots&D^{(m_{\mathrm{out}} - 1, m_{\mathrm{in}}-1)}\\
        \end{bmatrix}
      }. \qed
    \end{equation}

  \subsection{Proposition\,\ref{prop:stacked_conv}}

    We prove the proposition partially following \citet{Sedghi \etal}{SingularConv}.
    Using prop.\,\ref{prop:decomposition}, $M^{(i)}$ can be decomposed as follows.
    \begin{equation}
      M^{(i)} = {
        \left(I_{m_{i+1}}\otimes Q_N\right) L^{(i)} \left(I_{m_{i}}\otimes Q_N\right)^{\mathrm{H}}
      },
    \end{equation}
    where $L^(i)$ is a block matrix such that each block is diagonal.
    Since
    \begin{equation}
      \left(I_{m}\otimes Q_N\right)^{\mathrm{H}}\left(I_{m}\otimes Q_N\right)  = I_{mN^2},
    \end{equation}
    we can write $M$ as
    \begin{equation}
      M = {
        \left(I_{m_{d+1}}\otimes Q_N\right) \left(\prod_{i=1}^d L^{(i)}\right) \left(I_{m_{1}}\otimes Q_N\right)^{\mathrm{H}}
      },
    \end{equation}
    where $d$ is the number of layers.
    Let
    \begin{align}
      L &= \prod_{i=1}^d L^{(i)}\\
        &= {
        \begin{bmatrix}
          D^{(0,0)}&D^{(0,1)}&\dots&D^{(0, m_{\mathrm{in}}-1)}\\
          D^{(1,0)}&D^{(1,1)}&\dots&D^{(1, m_{\mathrm{in}}-1)}\\
          \vdots&\vdots&\ddots&\vdots\\
          D^{(m_{\mathrm{out}} - 1,0)}&D^{(m_{\mathrm{out}} - 1,1)}&\dots&D^{(m_{\mathrm{out}} - 1, m_{\mathrm{in}}-1)}\\
        \end{bmatrix}
      }.
    \end{align}
    Since all $L^{(i)}$ are block matrix such that all blocks are diagonal, $D^{(i,j)}$ are diagonal.
    For any $w\in\{1,\dots,N^2\}$, let $G^{(w)}$ be a matrix such that
    \begin{equation}
      G^{(w)}_{i,j} = D^{(i,j)}_{w,w}.
    \end{equation}
    Let $\sigma$ be a singular value of $G^{(w)}$ with a left singular vector $x$ and a right singular vector $y$.
    We claim that $y\otimes (Q_N)_{:,w}$ is a right singular vector of $M$.
    Let $e_w$ be a standard basis vector.
    Since $D^{(i,j)}$ is diagonal,
    \begin{align}
      L(y\otimes e_w) &= \sigma (x\otimes e_w).
    \end{align}
    Thus,
    \begin{align}
      M\left(y\otimes (Q_N)_{:,w}\right) &= (I_{m_{d+1}}\otimes Q_N)L(y\otimes e_w) \\
                                         &= \sigma (I_{m_{d+1}}\otimes Q_N) (x\otimes e_w) \\
                                         &= \sigma (x\otimes (Q_N)_{:,w}).
    \end{align}
    Let $\tilde{\sigma}$ be another singular value of $G^{(w)}$ with a left singular vector $\tilde{x}$ and a right singular vector $\tilde{y}$.
    Then,
    \begin{align}
      (x\otimes \left((Q_N)_{:,w}\right))^{\mathrm{H}}(\tilde{x}\otimes \left((Q_N)_{:,w}\right))
      &= (x\otimes \left((Q_N)_{:,w}\right))^{\mathrm{H}}(I_{m_{\mathrm{out}}}\otimes Q_N)(I_{m_{\mathrm{out}}}\otimes Q_N)^{\mathrm{H}}(\tilde{x}\otimes \left((Q_N)_{:,w}\right)) \\
      &= (x\otimes e_w)^{\mathrm{H}}(\tilde{x}\otimes e_w) \\
      &= x^{\mathrm{H}}\tilde{x} \\
      &= 0.
    \end{align}
    Similarly,
    \begin{align}
      (y\otimes \left((Q_N)_{:,w}\right))^{\mathrm{H}}(\tilde{y}\otimes \left((Q_N)_{:,w}\right)) &= 0.
    \end{align}
    Also,
    \begin{align}
      (x\otimes \left((Q_N)_{:,w}\right))^{\mathrm{H}}(x\otimes \left((Q_N)_{:,w}\right)) &= 1,\\
      (y\otimes \left((Q_N)_{:,w}\right))^{\mathrm{H}}(y\otimes \left((Q_N)_{:,w}\right)) &= 1.
    \end{align}
    Let $\tilde{\sigma}$ be another singular value of $G^{(\tilde{w})}$ with a left singular vector $\tilde{x}$ and a right singular vector $\tilde{y}$, where $w \neq \tilde{w}$.
    Then,
    \begin{align}
      (x\otimes \left((Q_N)_{:,w}\right))^{\mathrm{H}}(\tilde{x}\otimes \left((Q_N)_{:,\tilde{w}}\right))
      &= (x\otimes \left((Q_N)_{:,w}\right))^{\mathrm{H}}(I_{m_{\mathrm{out}}}\otimes Q_N)(I_{m_{\mathrm{out}}}\otimes Q_N)^{\mathrm{H}}(\tilde{x}\otimes \left((Q_N)_{:,\tilde{w}}\right)) \\
      &= (x\otimes e_w)^{\mathrm{H}}(\tilde{x}\otimes e_{\tilde{w}}) \\
      &= 0.
    \end{align}
    The last line holds because there are no overwrap in non-zero elements in the two vectors.
    Similarly,
    \begin{align}
      (y\otimes \left((Q_N)_{:,w}\right))^{\mathrm{H}}(\tilde{y}\otimes \left((Q_N)_{:,\tilde{w}}\right))
      &= 0.
    \end{align}
    Thus, using the Kronecker product of singular vectors of $G^{(w)}$ and $(Q_N)_{:,w}$ for all $w$,
    we may form a singular value decomposition of $M$. \qed

  \subsection{Proposition\,\ref{prop:skip_conv}}

    Let $M$ be a matrix that represents the convolutional layer.
    When we have a skip connection, the convolution plus the skip connection can be represented as
    \begin{align}
      M + I.
    \end{align}
    Since $M$ is a doubly block circulant matrix, $M+I$ is also a doubly block circulant matrix.
    Thus, we can apply Prop.\,\ref{prop:stacked_conv} with the number of layer $d=1$. \qed

  \subsection{Proposition\,\ref{prop:normalized_conv}}

    Normalization layers such as batch-normalization layer at test time or weight-normalization layer can be represented by a multiplication of a diagonal matrix whose elements corresponding to the same channels are equal.
    Thus, convolutional layers followed by such normalization layers can be represented by Eq.\,\eqref{eq:conv}.
    Thus, we can apply Prop.\,\ref{prop:stacked_conv}. \qed

  \subsection{Proposition\,\ref{prop:reduction}}

    First, we consider a sampling operation to a tensor $G$ such that we sample elements of inputs
    whose $x, y$ coordinates are in $\{i,j| i=0\ (\mathrm{mod}\ s) \text{ and } j=0\ (\mathrm{mod}\ s)\}$.
    For simplicity, we consider a convolution with intput output channel sizes are one.
    We start from analysis of the output of the operation when its input is $(F_N)_{:,a} \otimes (F_N)_{b,:}$.
    Since $i,j$-th element of the output is $\left((F_N)_{:,a} \otimes (F_N)_{b,:}\right)_{i\times s,j\times s}$ and
    \begin{equation}
      \left((F_N)_{:,a} \otimes (F_N)_{b,:}\right)_{i\times s,j\times s} = \left((F_{N/s})_{:,(a\%(N/s))} \otimes (F_{N/s})_{(b\%(N/s)),:}\right)_{i,j},
    \end{equation}
    the output is $(F_{N/s})_{:,(a\%(N/s))} \otimes (F_{N/s})_{(b\%(N/s)),:}$.
    Thus, when we decompose the input $x$ as
    \begin{equation}
      x = \sum_{a=0}^{N}\sum_{b=0}^{N} \lambda^{(a,b)} (F_N)_{:,a} \otimes (F_N)_{b,:},
    \end{equation}
    and decompose the output $y$ as
    \begin{equation}
      y = \sum_{a=0}^{N/s}\sum_{b=0}^{N/s} \tilde{\lambda}^{(a,b)} (F_N)_{:,a} \otimes (F_N)_{b,:},
    \end{equation}
    the following equation holds.
    \begin{equation}
      \tilde{\lambda}^{(a,b)} = \sum_{l=0}^{s}\sum_{r=0}^{s} \lambda^{(a + lN/s,b + rN/s)}.
    \end{equation}

    Let $M=(I_{\mathrm{m_{\mathrm{in}}}}\otimes Q)L(I_{\mathrm{m_{\mathrm{in}}}}\otimes Q)^H$ be a matrix that represents a convolutional layer with stride $1$.
    Since a convolutional layer with stride $s$ can be represented by a multiplication of $M$ followed by the sampling operation,
    it can be represented by $SM$.
    Now, we consider the singular value decomposition of $SM$.
    Let $\Lambda_w$ be a set of indices defined as follows.
    \begin{equation}
      \Lambda_w = \{w + lN^2/s + rN/s|l, r \in \{1, \dots, s\}\}.
    \end{equation}
    For any frequency $i$ and index $j\in\{1, \dots, m\}$,
    let $\sigma^{(i,j)}$ be a singular value of $M$ with a left singular vector $x^{(i,j)}\otimes (Q_N)_{:,i}$ and a right singular vector $y^{(i,j)}\otimes (Q_N)_{:,i}$.
    For any frequency $w$, $\{p^{(i,j)}|w\in \Lambda_w,j\in\{1,\dots,m_{\mathrm{in}}\}, p^{i,j} \in \mathbb{C}\}$, we have
    \begin{align}
      & SM\left(\underset{i\in\Lambda_w,j\in\{1,\dots,m_{\mathrm{in}}\}}{\sum}p^{(i,j)} y^{(i,j)}\otimes (Q_N)_{:,i}\right)\\
      =& S\left(\underset{i\in\Lambda_w,j\in\{1,\dots,m_{\mathrm{out}}\}}{\sum}\sigma^{(i,j)}p^{(i,j)} x^{(i,j)}\otimes (Q_N)_{:,i}\right)\\
      =& \left(\underset{i\in\Lambda_w,j\in\{1,\dots,m_{\mathrm{out}}\}}{\sum}\sigma^{(i,j)}p^{(i,j)}x^{(i,j)}\right)\otimes (Q_{N/s})_{:,w}.
    \end{align}
    Let $X$ be a $m_{\mathrm{out}}\times m_{\mathrm{in}}s^2$ matrix such that each column is $\sigma^{(i,j)}x^{(i,j)}$ and $X=U\Sigma V^H$ be a singular value decomposition of $X$.
    We claim that when we use a column of $V$ as $p$,
    \begin{equation}
    \label{eq:rsv}
      \underset{i\in\Lambda_w,j\in\{1,\dots,m_{\mathrm{in}}\}}{\sum}p^{(i,j)} y^{(i,j)}\otimes (Q_N)_{:,i}
    \end{equation}
    is a right singular vector of $SM$.

    Choose $p$ such that $Xp \neq 0$.
    Let $U_{p}$ be a column of $U$ corresponding to $p$, which is a colum of $V$.
    Let $\Sigma_{p}$ be a diagonal element of $\Sigma$ corresponding to $p$.
    We have
    \begin{align}
      & SM\left(\underset{i\in\Lambda_w,j\in\{1,\dots,m_{\mathrm{in}}\}}{\sum}p^{(i,j)} y^{(i,j)}\otimes (Q_N)_{:,i}\right)\\
      =& \left(\underset{i\in\Lambda_w,j\in\{1,\dots,m_{\mathrm{out}}\}}{\sum}\sigma^{(i,j)}p^{(i,j)}x^{(i,j)}\right)\otimes (Q_{N/s})_{:,w}\\
      =& \Sigma_pU_p\otimes (Q_{N/s})_{:,w}.
    \end{align}

    For any $p$, since all $y^{(i,j)} \otimes (Q_N)_{:,i}$ are linearly independent, we have
    \begin{align}
      &\left(\underset{i\in\Lambda_w,j\in\{1,\dots,m\}}{\sum}p^{(i,j)} y^{(i,j)}\otimes (Q_N)_{:,i}\right)^{H}\left(\underset{i\in\Lambda_w,j\in\{1,\dots,m\}}{\sum}p^{(i,j)} y^{(i,j)}\otimes (Q_N)_{:,i}\right)\\
      =& \underset{i\in\Lambda_w,j\in\{1,\dots,m\}}{\sum}\|p^{(i,j)}\|^2\\
      =& 1.
    \end{align}
    Also, when $\tilde{p}$ is a different column of $V$,
    \begin{align}
      &\left(\underset{i\in\Lambda_w,j\in\{1,\dots,m\}}{\sum}p^{(i,j)} y^{(i,j)}\otimes (Q_N)_{:,i}\right)^{H}\left(\underset{i\in\Lambda_w,j\in\{1,\dots,m\}}{\sum}\tilde{p}^{(i,j)} y^{(i,j)}\otimes (Q_N)_{:,i}\right)\\
      =& \underset{i\in\Lambda_w,j\in\{1,\dots,m\}}{\sum}(p^{(i,j)})^{*}\tilde{p}^{(i,j)}\\
      =& 0.
    \end{align}
    Let $\tilde{w}\neq w$ be a different frequency and
    \begin{equation}
      \underset{i\in\Lambda_{\tilde{w}},j\in\{1,\dots,m\}}{\sum}\tilde{p}^{(i,j)} \tilde{y}^{(i,j)}\otimes (Q_N)_{:,i}
    \end{equation}
    be a vector constructed with the same way as \eqref{eq:rsv} for $\tilde{w}$.
    Since all vectors in the union of $\{y^{(i,j)}\otimes (Q_N)_{:,i}| i\in \Lambda_w,j\in\{1,\dots,m\}\}$ and $\{\tilde{y}^{(i,j)}\otimes (Q_N)_{:,i}|i\in\Lambda_{\tilde{w}},j\in\{1,\dots,m\}\}$ are linearly independent,
    \begin{align}
      \left(\underset{i\in\Lambda_w,j\in\{1,\dots,m\}}{\sum}p^{(i,j)} y^{(i,j)}\otimes (Q_N)_{:,i}\right)^{H}\left(\underset{i\in\Lambda_{\tilde{w}},j\in\{1,\dots,m\}}{\sum}\tilde{p}^{(i,j)} \tilde{y}^{(i,j)}\otimes (Q_N)_{:,i}\right)
      = 0.
    \end{align}

    Since $U$ is unitary, for any column $U_p$,
    \begin{align}
      (U_p\otimes (Q_{N/s})_{:,w})^H(U_p\otimes (Q_{N/s})_{:,w}) = 1,
    \end{align}
    and
    \begin{align}
      (U_p\otimes (Q_{N/s})_{:,w})^H(U_{\tilde{p}}\otimes (Q_{N/s})_{:,w}) = 0
    \end{align}
    when $\tilde{p}\neq p$.
    Let $\tilde{w}\neq w$ be a different frequency. Since $(Q_{N/s})_{:,w}$ and $(Q_{N/s})_{:,\tilde{w}}$ are linearly independent,
    \begin{align}
      (U_p\otimes (Q_{N/s})_{:,w})^H(\tilde{U}_{\tilde{p}}\otimes (Q_{N/s})_{:,\tilde{w}}) = 0.
    \end{align}

    Thus, for any frequency $w$, $\Sigma_p$ is a singular value of $SM$ with a left singular vector $U_p\otimes (Q_{N/s})_{:,w}$ and a right singular vector
    \begin{equation}
      \underset{i\in\Lambda_w,j\in\{1,\dots,m\}}{\sum}p^{(i,j)} y^{(i,j)}\otimes (Q_N)_{:,i}.
    \end{equation}
    We may form a singular value decomposition of $SM$ by using them. \qed

  \subsection{Proposition\,\ref{prop:complex}}

    Assume $x$ is a vector such that $S(x)_{u,v} = S(x)^{*}_{N-u,N-v}$.
    Let $y = S(x)$. Then,
    \begin{align}
      x_{u,v} &= S^{-1}(y)_{u,v} \\
              &= \frac{1}{N}\underset{m=0}{\overset{N-1}{\sum}}
                \underset{n=0}{\overset{N-1}{\sum}}
                y_{m,n}\exp(2\pi \sqrt{-1}(um + vn)/N) \\
              &= \frac{1}{2N}\underset{m=0}{\overset{N-1}{\sum}}
                \underset{n=0}{\overset{N-1}{\sum}}
                \big(\nonumber\\&y_{m,n}\exp(2\pi \sqrt{-1}(um + vn)/N) +y_{N-m,N-n}\exp(2\pi \sqrt{-1}(u(N-m) + v(N-n))/N)\big) \\
              &= \frac{1}{N}\underset{m=0}{\overset{N-1}{\sum}}
                \underset{n=0}{\overset{N-1}{\sum}}
                \left(\mathrm{Re}(y_{m,n})\cos(2\pi (um + vn)/N) - \mathrm{Im}(y_{m,n})\sin(2\pi (um + vn)/N)\right),
    \end{align}
    which is real.
    Thus, when $S(x)_{u,v} = S(x)^{*}_{N-u,N-v}$ for all $u$ and $v$, $x$ is real.

    Assume $x$ is a real vector.
    Then,
    \begin{align}
      S(x)_{N-u,N-v}^{*} &= \frac{1}{N}\underset{m=0}{\overset{N-1}{\sum}}
                         \underset{n=0}{\overset{N-1}{\sum}}
                         y_{m,n}\exp(2\pi \sqrt{-1}((N-u)m + (N-v)n)/N)^{*} \\
                      &= \frac{1}{N}\underset{m=0}{\overset{N-1}{\sum}}
                         \underset{n=0}{\overset{N-1}{\sum}}
                         y_{m,n}\exp(-2\pi \sqrt{-1}((N-u)m + (N-v)n)/N) \\
                      &= \frac{1}{N}\underset{m=0}{\overset{N-1}{\sum}}
                         \underset{n=0}{\overset{N-1}{\sum}}
                         y_{m,n}\exp(2\pi \sqrt{-1}(um + vn)/N) \\
                      &= S(x)_{u,v}. \qed
    \end{align}

\section{Evaluation setups}

  \paragraph{Datasets:}
    We used MNIST\,\citep{MNIST}, fashion-MNIST\,\citep{FMNIST}, SVHN\,\citep{SVHN}, CIFAR10, CIFAR100\,\citep{CIFAR}, and ILSVRC2015\,\citep{ILSVRC} as datasets.
    For CIFAR10 and CIFAR100, as an data augmentation, we padded four pixels on each side and randomly sampled a $32\times 32$ crop from the padded image or its horizontal flip.
    4 pixels are padded on each side
    We then normalized them with the mean and std of each channel.
    For training on ILSVRC2015, we augmented data following \citet{He \etal}{ResNet}.
    For training on ILSVRC2015, we rescaled images with its shorter side randomly sampled in $[256, 480]$ and randomly cropped into $224\times244$ for scale augmentation\,\citep{VGG}. We used per-channel subtraction and standard color augmentation\,\citep{AlexNet}.
    For other datasets, we scaled inputs into the range from zero to one.

  \paragraph{Architectures:}
    We used a multi-layer perceptron (MLP) consisting of $1000$--$1000$ hidden layer with ReLU activation,
    LeNet\,\citep{LeNet}, WideResNet\,\citep{WideResNet}, DenseNet-BC\,\citep{DenseNet}, and VGG\,\citep{VGG} with batch-normalization for evaluations on datasets except for ILSVRC2015.
    For ILSVRC2015, we used ResNet50\,\citep{ResNet}, DenseNet, VGG16, and GoogLeNet\,\citep{GoogLeNet}.
    For VGG16 and GoogLeNet, we added a batch-normalization layer after each convolution for faster training.

  \paragraph{Training details except for ILSVRC2015:}
    We used Nesterov momentum as an optimizer with momentum $0.9$, weight decay $0.0005$, and batchsize $128$ for the experiments.
    We trained the MLP and LeNet for 50 epochs with an initial learning rate $0.1$ decayed by $0.1$ at every $10$ epochs.
    We trained WideResNet as follows.
    For MNIST, fashion-MNIST, and SVHN, we used width factor $k=4$, layer $16$, and dropout ratio $0.4$, and trained for $160$ epochs with initial learning ratio $0.01$ decayed by $0.1$ at epoch $80$ and $120$.
    For CIFAR10 and CIFAR100, we used width factor $k=10$, layer $28$, and dropout ratio $0.3$, and trained for $200$ epochs with initial learning ratio $0.1$ decayed by $0.1$ at epoch $60$, $120$, and $160$.
    These are the same configuration for SVHN and CIFAR in \citet{Zagoruyko and Komodakis}{WideResNet}.
    We trained DenseNet-BC with layer $100$, growth rate $12$, and dropout ratio $0.2$.

  \paragraph{Training details on ILSVRC2015:}
    We used SGD with momentum $0.9$, weight decay $0.0001$, and batchsize $256$, and trained for $90$ epochs for all architectures.
    For ResNet50, GoogLeNet, and VGG16, we used the same learning rate scheduling and momentum correction used by \citet{Goyal \etal}{OneHour}.
    For DenseNet121, we set an initial learning rate to $0.1$ and multiplied by $0.1$ at epoch $30$ and $60$, following \citet{Huang \etal}{DenseNet}.

  \paragraph{Metric:}
    We used the fool ratio as a metric, which is the percentage of data that models changed its prediction, following \citet{Moosavi{-}Dezfooli \etal}{Universal}.

\end{document}